%% file: main.tex
\documentclass[10pt,journal, compsoc]{IEEEtran}
\usepackage[nocompress]{cite} 
\usepackage[bookmarks,backref=true,linkcolor=black]{hyperref} 
\usepackage{verbatim} 
\usepackage{ulem} 
\usepackage{makecell}
\usepackage{mathptmx}
\usepackage[pdftex]{graphicx}
\usepackage{epstopdf}
\usepackage{times}
\usepackage[ruled,vlined]{algorithm2e}  
\usepackage{amssymb,amsmath}
\usepackage{multirow}
\usepackage{subfigure}
\usepackage{paralist}
\usepackage{tikz}
\usepackage{multirow}
\usepackage{color}
\usepackage{ulem}
\usepackage{setspace}
\usepackage{microtype,hyphenat,balance}
\usepackage{array}
\usepackage{url}
\usepackage{bm}
\usepackage{overpic}
\usepackage{contour}

\usepackage{amsfonts}
\usepackage{bbm}
\usepackage{diagbox}
\usepackage{url}
\usepackage{caption}
\usepackage{wrapfig}
\usepackage{microtype,hyphenat,balance,booktabs}

\usepackage{ragged2e}

\DeclareMathAlphabet{\mathcal}{OMS}{cmsy}{m}{n}

\newcommand{\Fig}[1]{Fig.~\ref{#1}}
\newcommand{\Eq}[1]{Eq.~(\ref{#1})}
\newcommand{\Sec}[1]{Sec.~\ref{#1}}

\newcommand{\E}[1]{{E{#1}}}

\newcommand{\yang}[1]{\textcolor{black}{#1}}
\newcommand{\liu}[1]{\textcolor{black}{#1}}
\newcommand{\xia}[1]{\textcolor{black}{#1}}
\newcommand{\changjian}[1]{\textcolor{black}{#1}}
\newcommand{\minor}[1]{\textcolor{black}{#1}}

\def \etal {{\emph{et al}.}}

\def \eg {{\emph{e.g}.}}

\ifCLASSOPTIONcompsoc
  \usepackage[nocompress]{cite}
\else
  \usepackage{cite}
\fi
\ifCLASSINFOpdf
\else
\fi
\begin{document}
%
\title{Diagnosing Ensemble Few-Shot Classifiers}
%
%
%
%

\author{Weikai Yang, Xi Ye, Xingxing Zhang, Lanxi Xiao, Jiazhi Xia, Zhongyuan Wang, Jun Zhu, Hanspeter Pfister, and Shixia Liu
\IEEEcompsocitemizethanks{
\IEEEcompsocthanksitem W.~Yang, X.~Zhang, L.~Xiao, J.~Zhu, and S.~Liu are with Tsinghua University. 
\IEEEcompsocthanksitem X.~Ye is with the University of Texas at Austin.
\IEEEcompsocthanksitem J.~Xia is with Central South University.
\IEEEcompsocthanksitem Z.~Wang is with Kuaishou Technology Co., Ltd.
\IEEEcompsocthanksitem H.~Pfister is with Harvard University.
}
}

\input{0.Abstract.tex}


\maketitle
\IEEEdisplaynontitleabstractindextext
\IEEEpeerreviewmaketitle

{
\fontsize{10}{10} 
\input{1.Introduction}

\input{2.RelatedWork}

\input{3.Background}

\input{4.Requirements}

\input{5.Algorithm}

\input{6.Visualization}
\input{7.Evaluation}

\input{8.Discussion}
\input{9.Conclusion}
}

\ifCLASSOPTIONcompsoc
 \section*{Acknowledgments}
\else
 \section*{Acknowledgment}
\fi
This work was supported by the National Key R\&D Program of China under Grant 2020YFB2104100, the National Natural Science Foundation of China under grants U21A20469, 61936002, grants from the Institute Guo Qiang, THUIBCS, and BLBCI, and in part by Tsinghua-Kuaishou Institute of Future Media Data.

\ifCLASSOPTIONcaptionsoff
  \newpage
\fi

\bibliographystyle{IEEEtran}
\bibliography{main}
\vspace{-12mm}

\begin{IEEEbiography}
[{\includegraphics[width=1in,height=1.25in,clip,keepaspectratio]{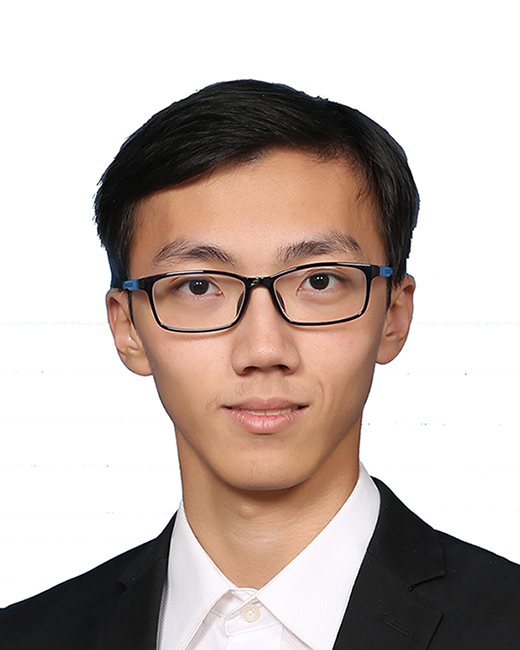}}]
{Weikai Yang} is a Ph.D. student at Tsinghua University. His research interests include visual text analytics and interactive machine learning. He received a B.S. degree from Tsinghua University.
\end{IEEEbiography}

\begin{IEEEbiography}
[{\includegraphics[width=1in,height=1.25in,clip,keepaspectratio]{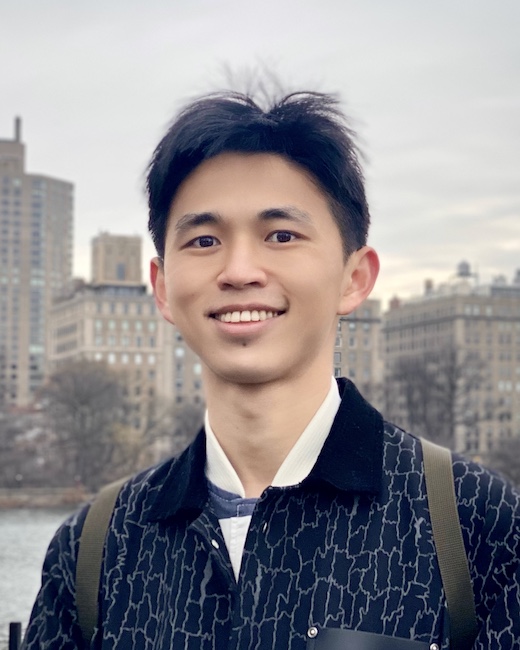}}]
{Xi Ye} is a Ph.D. student at the University of Texas at Austin. Xi received a B.S. degree from Tsinghua University. His research focuses on building interpretable and robust models for complex NLP tasks.
\end{IEEEbiography}

\begin{IEEEbiography}[{\includegraphics[width=1in,height=1.25in,clip,keepaspectratio]{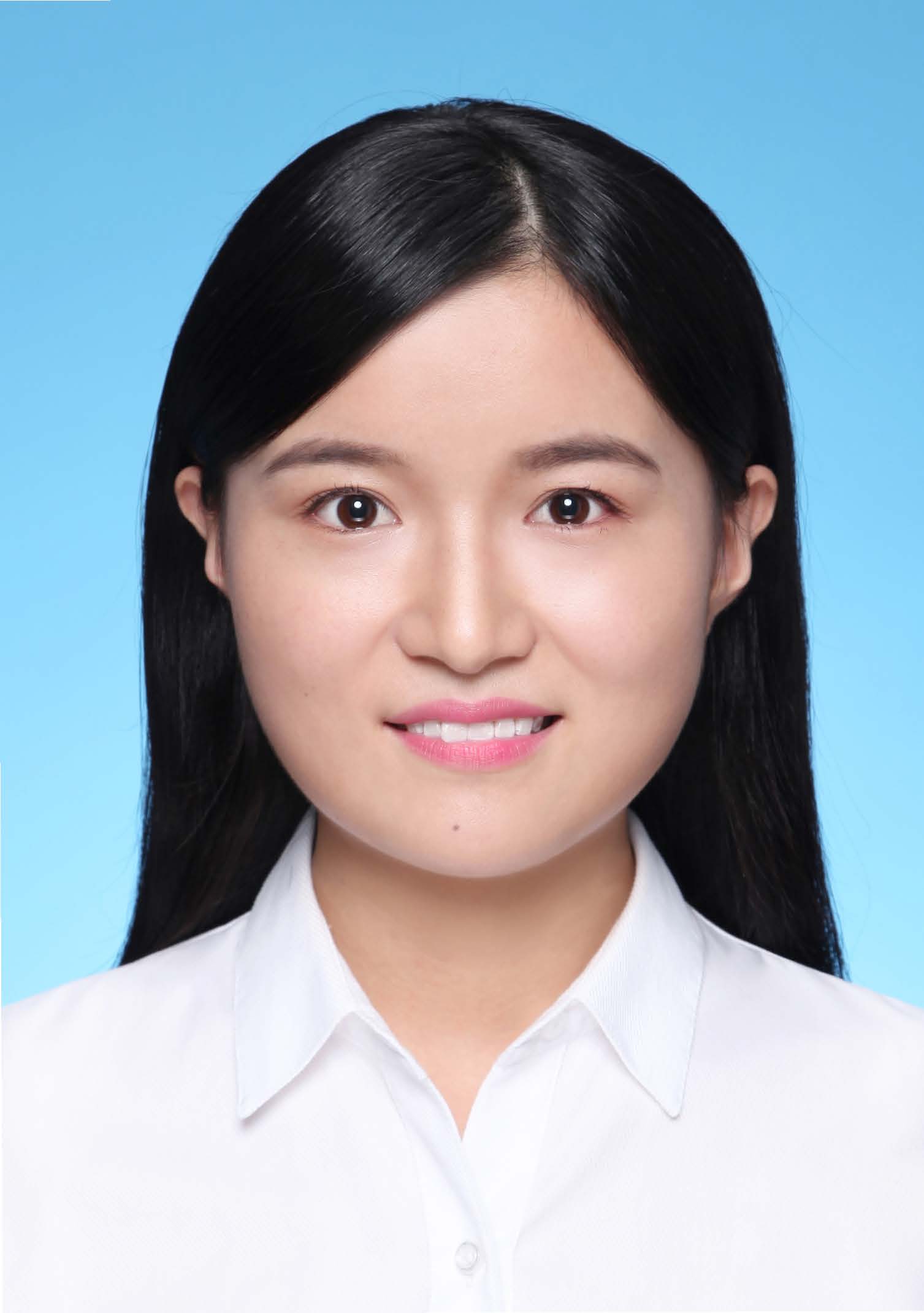}}]{Xingxing Zhang} is a Postdoc in the department of computer science, Tsinghua University. Her research interests include data selection, zero/few-shot learning, and adversarial learning. She received a B.E. degree and Ph.D. degree in signal and information processing from Beijing Jiaotong University in 2015 and 2020, respectively. She received the excellent Ph.D. thesis award from the Chinese Institute of Electronics in 2020.
\end{IEEEbiography}

\begin{IEEEbiography}
[{\includegraphics[width=1in,height=1.25in,clip,keepaspectratio]{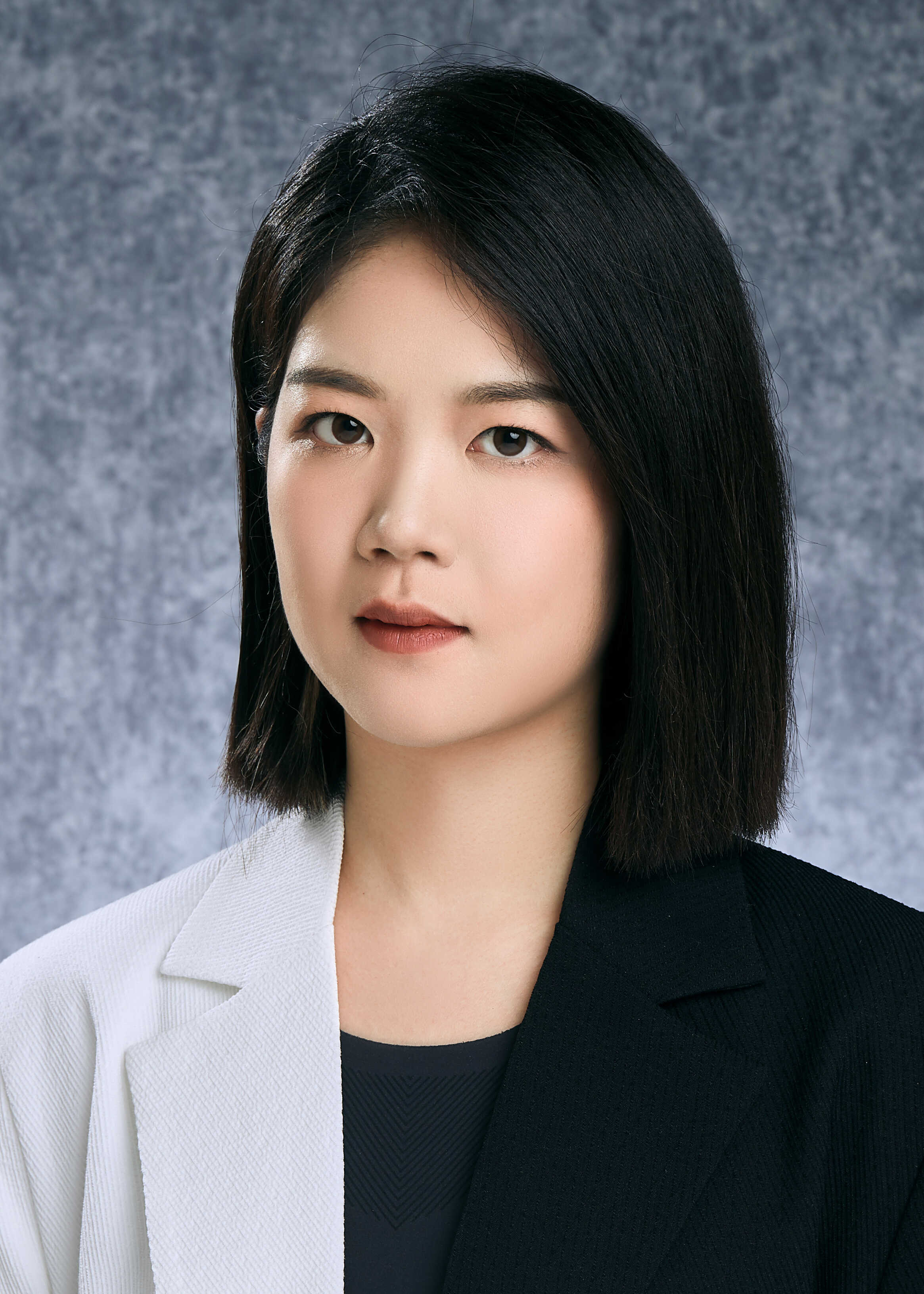}}]
{Lanxi Xiao} is a master student at Tsinghua University. Her research interests include information, interaction, and innovation design, as well as Human-AI collaboration design. She received a B.A. degree in Art \& Technology (Information Design) from Tsinghua University in 2020. 
\end{IEEEbiography}

\begin{IEEEbiography}
[{\includegraphics[width=1in,height=1.25in,clip,keepaspectratio]{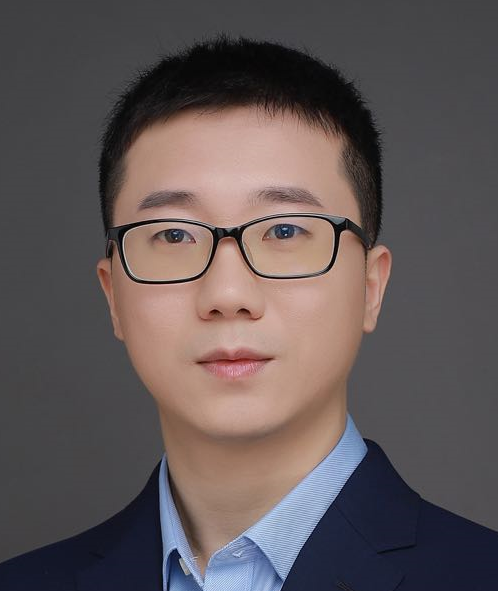}}]
{Jiazhi Xia} is a professor in the School of Computer Science and Engineering at Central South University. His research interest includes data visualization, visual analytics, and computer graphics. Recently, he has performed research in high-dimensional data visualization, graph visualization, and visual analytics for machine learning and published more than 10 IEEE VIS papers. He has served as the survey paper co-chair of ChinaVis in 2019-2020, paper co-chair of ChinaVis in 2021.
\end{IEEEbiography}

\begin{IEEEbiography}
[{\includegraphics[width=1in,height=1.25in,clip,keepaspectratio]{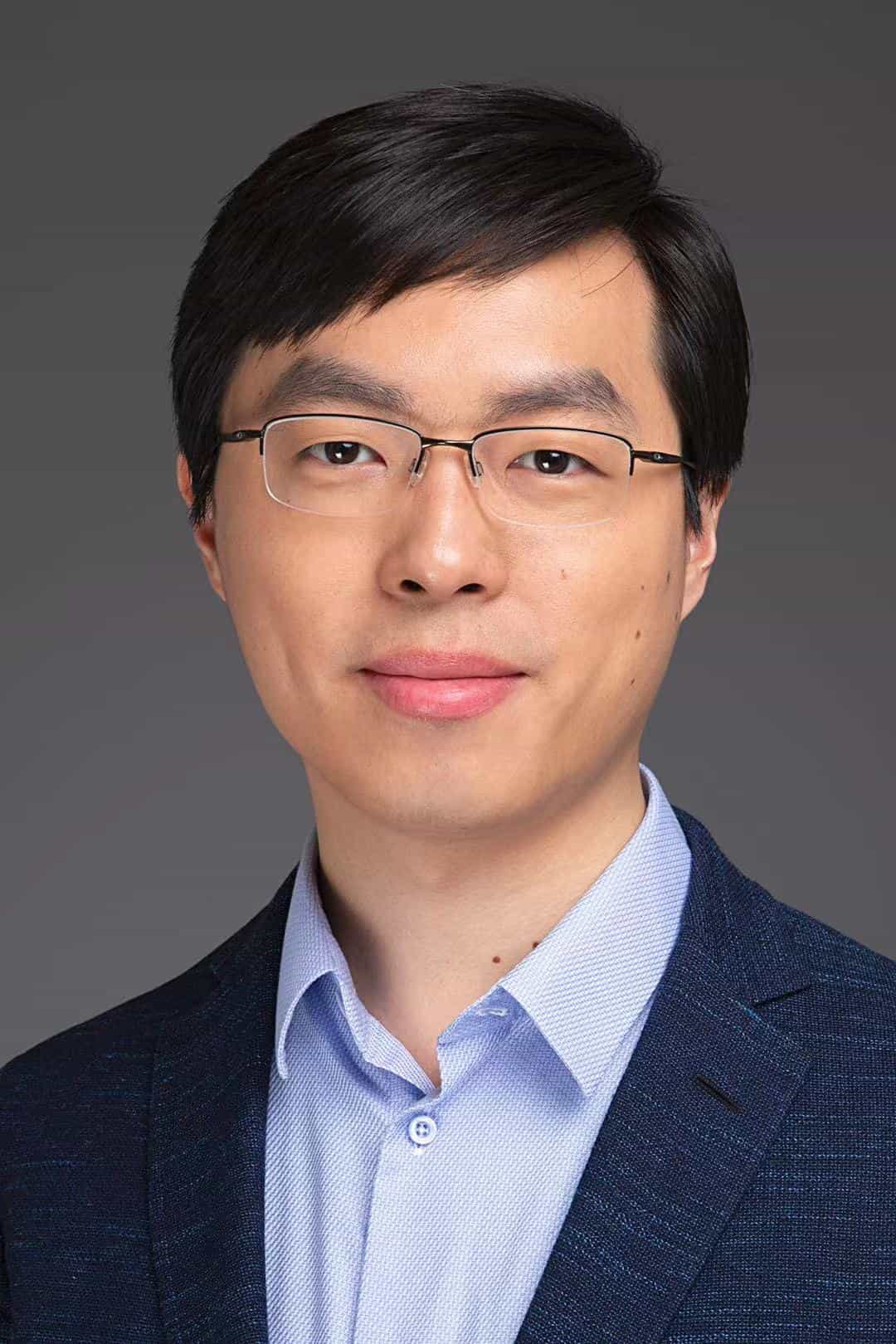}}]
{Zhongyuan Wang} is the director of the AI team and a vice president of Kuaishou. His research interests include knowledge graph, natural language processing, information retrieval, and deep learning. Before Kuaishou, he worked for Meituan, Facebook, and Microsoft Research separately and led the research and development of artificial intelligence techniques. He received a B.S., an M.S., and a Ph.D. from Renmin University of China. He received the 2018 MIT TR Innovators Under 35 China.
\end{IEEEbiography}

\begin{IEEEbiography}
[{\includegraphics[width=1in,height=1.25in,clip,keepaspectratio]{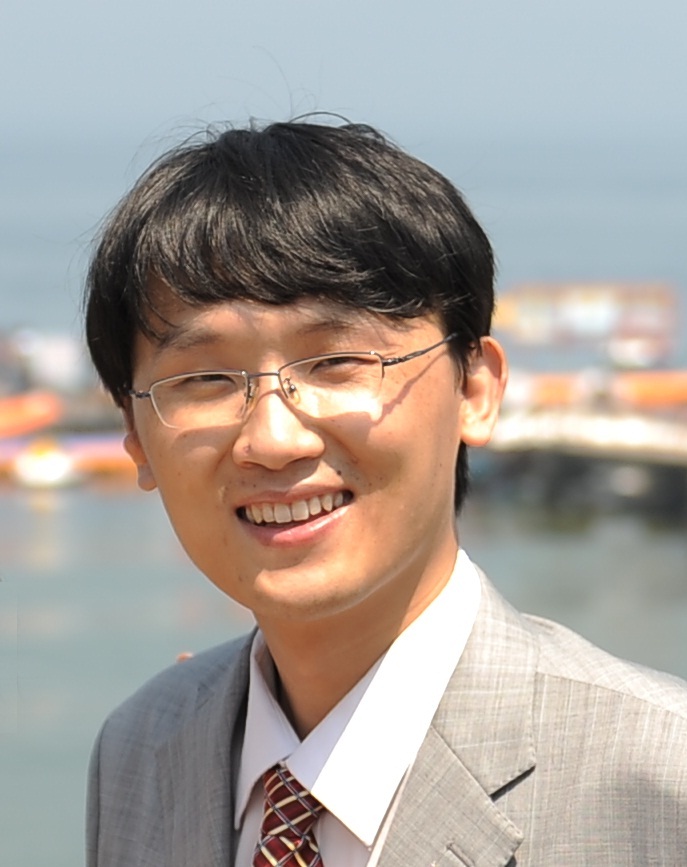}}]
{Jun Zhu} is a professor at Tsinghua University. His research interests include primarily on developing statistical machine learning methods to understand scientific and engineering data arising from various fields. He received the BS and PhD degrees from the Department of Computer Science and Technology, Tsinghua University. He was an adjunct faculty of Carnegie Mellon University. He is an associate editor-in-chief of IEEE TPAMI.
\end{IEEEbiography}

\begin{IEEEbiography}
[{\includegraphics[width=1in,height=1.25in,clip,keepaspectratio]{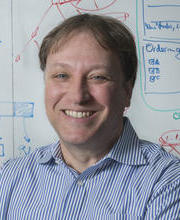}}]
{Hanspeter Pfister} is An Wang Professor at Harvard University. His research interest includes biomedical image analysis and visualization, image and video analysis, and visual analytics in data science. He worked as an Associate Director and Senior Research Scientist at Mitsubishi Electric Research Laboratories. He has a Ph.D. from the State University of New York at Stony Brook and an M.S. in Electrical Engineering from ETH Zurich. He is the recipient of the 2010 IEEE Visualization Technical Achievement award.
\end{IEEEbiography}

\begin{IEEEbiography}
[{\includegraphics[width=1in,height=1.25in,clip,keepaspectratio]{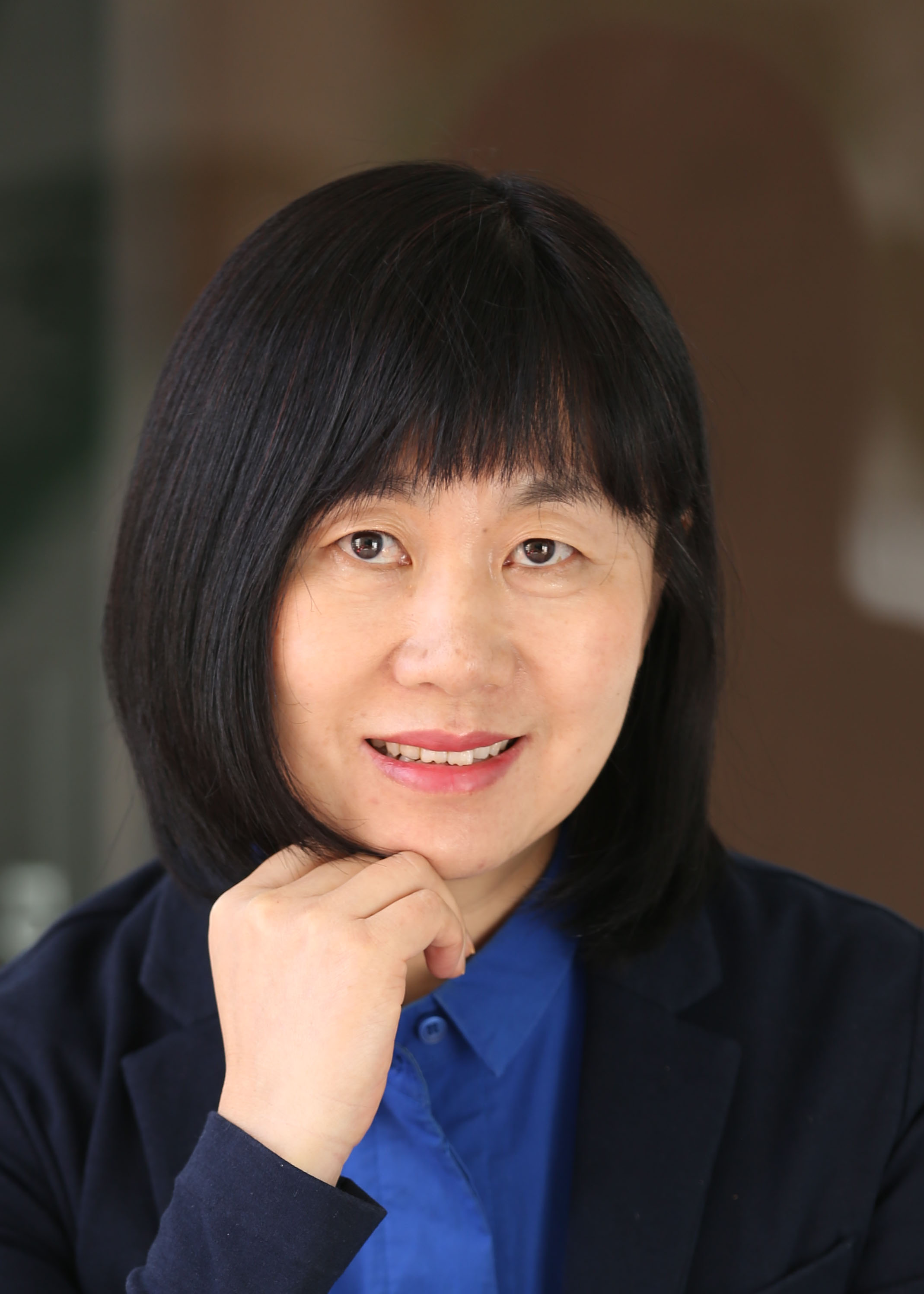}}]
{Shixia Liu}
is a professor at Tsinghua University. Her research interests include visual text analytics, visual social analytics, interactive machine learning, and text mining. She worked as a research staff member at IBM China Research Lab and a lead researcher at Microsoft Research Asia.
She received a B.S. and M.S. from Harbin Institute of Technology, a Ph.D. from Tsinghua University.
She is a fellow of IEEE and an associate editor-in-chief of IEEE Trans. Vis. Comput. Graph.
\end{IEEEbiography}





\end{document}

%% file: 0.Abstract.tex
\IEEEtitleabstractindextext{%
\justify
\begin{abstract}
The base learners and labeled samples (shots) in an ensemble few-shot classifier greatly affect the model performance.
When the performance is not satisfactory, it is usually difficult to understand the underlying causes and make improvements.
To tackle this issue, we propose a visual analysis method, FSLDiagnotor.
Given a set of base learners and a collection of samples with a few shots, 
we consider two problems: 1) finding a subset of base learners that well predict the sample collections; and 2) replacing the low-quality shots with more representative ones to adequately represent the sample collections.
We formulate both problems as sparse subset selection and develop two selection algorithms to recommend appropriate learners and shots, respectively.
A matrix visualization and a scatterplot are combined to explain the recommended learners and shots in context and facilitate users in adjusting them. 
Based on the adjustment, the algorithm updates the recommendation results for another round of improvement.
Two case studies are conducted to demonstrate that FSLDiagnotor helps build a few-shot classifier efficiently and increases the accuracy by 12\% and 21\%, respectively.
\end{abstract}

\begin{IEEEkeywords}
Few-shot learning, ensemble model, subset selection, matrix visualization, scatterplot
\end{IEEEkeywords}}

%% file: 1.Introduction.tex
\section{Introduction}
\IEEEPARstart{T}{he} few-shot classification aims to train a classifier to recognize unseen classes with only a few labeled samples (shots) in each class, which is of great significance both academically and practically~\cite{ dvornik2019diversity,wang2020generalizing}.
\xia{
For example, \minor{at the early stage of the COVID-19 epidemic,} the massive labeling of the CT scans requires a long process of clinical observation with the risk to patients' lives.}
As such, few-shot classification \liu{is a viable choice for} these scenarios.
\liu{Many} advances have been made to continuously improve the performance of few-shot classifiers by developing a variety of methods, such as ensemble learning, generative models, and meta-learning~\cite{wang2020generalizing}.
\liu{Because the ensemble few-shot classification can combine any few-shot classifiers (base learners) for better performance, it is the most widely used state-of-the-art method in practice.} 
For example, \liu{three of the top five best-performing models in \minor{a CVPR challenge on few-shot learning}~\cite{cvpr_challenge} and four of the top five best-performing models in \minor{a Kaggle competition on few-shot learning}~\cite{kaggle_competition} have used ensemble few-shot classifiers to boost performance successfully.}

Previous studies have shown that the performance of the ensemble model is largely affected by the diversity and cooperation among the individual base learners~\cite{dvornik2019diversity} and the representativeness of the shots~\cite{liu2018interactive}.
\changjian{
Accordingly, using all learners and shots may downgrade the performance. 
For example, if the performance of a learner is poor and its predictions are different from the majority, it will hurt the performance of the ensemble model.
In addition, a shot wrongly representing some samples usually leads to the misclassification of these samples.}
\liu{Thus, it is desirable to select a subset of cooperative and diverse learners and identify a small set of representative shots, which} is a long-standing challenge for the practical application of few-shot classification.
\liu{Existing learning methods typically apply an ensemble model to all the given learners and shots~\cite{dvornik2019diversity, qi2020few}, which often fail to achieve the best performance.
Improving the performance usually requires repeatedly selecting the learners and adjusting their weights.}
Without a comprehensive understanding of how the model and shots work together to reach the final predictions, this trial-and-error process is very time-consuming and expertise-demanding.
\liu{Moreover, lacking the refinement of the shots, the performance improvement is limited~\cite{renggli2021a,ng2021mlops}}.
To improve the performance efficiently, users need \yang{an efficient} way to analyze the performance-related log data (``\textbf{analyze first}").
The learners and shots with unusual behavior, such as the learner causing a large confidence drop or the shot with poor coverage, can be highlighted (``\textbf{show the important}").
\liu{After understanding the roles of learners/shots in the final predictions,}
they can then decide which ones to be added/removed for improving the performance (``\textbf{interaction and feedback}").
Based on the updated learners (shots), suitable shots (learners) are recommended for another round of analysis (``\textbf{analyze again}").
Such an iterative analysis process with human-in-the-loop fits well with the visual analytics mantra~\cite{keim2008visual} and inspires us to develop a visual analysis tool, FSLDiagnotor, for tuning the selection of learners and shots.

The key behind FSLDiagnotor is its ability to efficiently identify and eliminate performance bottlenecks caused by the selected base learners and shots.
Given a set of learners and a collection of samples with a few shots, 
we consider two problems:
1) finding a subset of diverse and cooperative learners that well predict the sample collections and
2) removing low-quality shots and recommending necessary new shots to adequately represent the sample collection.
By studying the intrinsic characteristics of these two problems, we formulate them as sparse subset selection and develop two selection algorithms to recommend appropriate learners and shots. 
However,
the recommendations are not always perfect and may contain one or a few low-quality learners/shots.
For example, a \yang{learner that wrongly predicts some samples with high confidence can be recommended because it is mistaken as a well-performing learner for those samples}.
Such low-quality learners/shots are hard to be detected and corrected without human involvement.
To facilitate such tasks, a matrix visualization and a scatterplot are combined to explain the prediction behavior of the recommended learners and the coverage of the shots in context. 
\yang{Based on the understanding of the behavior of the learners and shots}, users can improve the selection of learners and enhance the shots for better performance.\looseness=-1

We performed a quantitative evaluation to show that both the learner and shot selection algorithms can boost the performance of the few-shot classifier.
We also conducted two case studies with two machine learning experts to demonstrate that our tool helps diagnose and improve the few-shot classifier more efficiently and increases the accuracy by 12\% and 21\%, respectively.
The demo is available at \href{http://fsldiagnotor.thuvis.org/}{\textcolor{black}{http://fsldiagnotor.thuvis.org/}}.

The main contributions of this work include: 
\begin{compactitem}

\item\noindent{The formulation of sparse subset selection that unifies the shot and learner selection into one framework.}

\item\noindent{An enhanced matrix visualization coordinated with a scatterplot to explain how the base learners and shots contribute to the final predictions.}

\item\noindent{A visual analysis pipeline that tightly integrates the subset selection algorithm with interactive visualization to facilitate the iterative improvement of the shots and base learners.}

\end{compactitem}

%% file: 2.RelatedWork.tex
\section{Related Work}

\subsection{Few-Shot Classification}
\label{sec:relatedfsl}
The ensemble methods have been explored in the vein of few-shot classification to boost the performance~\cite{dvornik2019diversity,qi2020few}.
Dvornik~\etal\cite{dvornik2019diversity} encouraged the diversity and cooperation between learners for better performance.
In addition to training the base learners, 
Qi~\etal\cite{qi2020few} adaptively assigned a weight to each learner for a strong few-shot classifier.
While more and more sophisticated models have been developed, there is recent work pointing out the cruciality of high-quality features: using high-quality features is even more effective than employing a well-designed complex model~\cite{tian2020rethink}.
Following such a philosophy, 
Dvornik~\etal\cite{dvornik2020selecting} learned high-quality feature extractors to extract high-quality features for unlabeled samples.
Due to the importance of the diversity-cooperation strategy and the features,
our work combines the two.
We leveraged deep learning models, such as a pre-trained ResNet model~\cite{He2016CVPR}, to extract the features for each sample.
Then a set of learners were built based on the extracted features.
This saves training time and provides the flexibility to quickly obtain the base learners.
Our method also recommends a subset of base learners and enhances the quality of shots to further improve the performance. \looseness=-1

\subsection{Visual Analysis for Improving Model Performance}
Existing visual analysis work for improving model performance can be classified into two categories: model-driven methods and data-driven methods~\cite{hohman2018visual,yuan2021survey}. 

Model-driven methods facilitate experts to better understand 
the inner workings of a machine learning model and discover the reason why a training process fails to achieve an acceptable performance. 
For example, CNNVis~\cite{liu2016towards} was developed to diagnose the potential issues of a convolutional neural network (CNN) by examining the learned features and activation of neurons.
Alsallakh~\etal\cite{bilal2017convolutional} utilized a confusion matrix to disclose the impact of class hierarchy on the features learned at each CNN layer. 
Kahng~\cite{kahng2017activis} developed ACTIVIS to facilitate the identification of specific training issues on an industry-scale deep learning model by illustrating how neurons are activated by the instances of interest.
Later efforts focus on diagnosing other types of models, such as deep generative models~\cite{liu2017analyzing}, Deep Q-Networks~\cite{wang2018dqnviz}, and sequential models~\cite{ming2017understanding,strobelt2018seq}.
In addition to improving a single model, some efforts focus on analyzing ensemble models~\cite{Liu2018Visual, Schneider2021,Zhao2019,Neto2021}.
For example, Schneider~\etal\cite{Schneider2021} developed a visual analysis tool to explore the data and model spaces of the ensemble model and improve its performance by enabling a selection of the base learners.
Our method supports the improvement on both the data and model.

\liu{In the same spirit of data-centric AI~\cite{renggli2021a,ng2021mlops}, data-driven methods aim to} improve the quality of training samples at the instance and label levels.
At the instance level, Chen~\etal\cite{chen2020oodanalyzer} developed OoDAnalyzer, a visual analysis tool to analyze the out-of-distribution samples in the context of the training and test samples.
\yang{Yang~\etal\cite{yang2020diagnosing} proposed DriftVis to detect and correct the distribution changes in a data stream.}
Ming~\etal\cite{ming2020protosteer} developed ProtoSteer to explain the prediction of an input sample by using exemplary samples that have similar scores to this sample.
Model developers can improve the model performance by revising the exemplary samples. 
More recently, Gou~\etal\cite{gou2021vatld} proposed to generate unseen test cases to improve model robustness. 
At the label level, Heimerl~\etal\cite{heimerl2012visual} utilized active learning to facilitate the task of interactive labeling for document classification.
This idea of employing active learning to support interactive labeling has also been adopted by other visual analysis work~\cite{behrisch2014feedback,bruneau2013interactive,hoferlin2012inter,paiva2015approach}.
Most of the later research along this line has focused on detecting and correcting noisy labels in training samples. 
Liu~\etal\cite{liu2018interactive} introduced LabelInspect to improve the crowdsourced annotations by utilizing the mutual reinforcement relationships between the workers' behavior and the uncertainty of the annotated results.
Xiang~\etal\cite{xiang2019interactive} developed a visual analysis tool to correct label errors in a large set of training samples based on user-selected trust items.
More recently, Jia~\etal\cite{jia2021towards} applied active learning to zero-shot classification.
\liu{They interactively built a class attribute matrix for improving the performance of classifiers.}

Although the aforementioned methods have shown the capability of improving the model performance to some extent, there are few efforts that tightly combine model-driven methods with data-driven methods to improve performance.
The combination is particularly needed in few-shot learning since both the data and model greatly influence the performance. 
Thus, we develop FSLDiagnotor to improve both shots and learners.\looseness=-1

%% file: 3.Background.tex
\section{Background}

\begin{figure}[!b]
\centering
{\includegraphics[width=\linewidth]{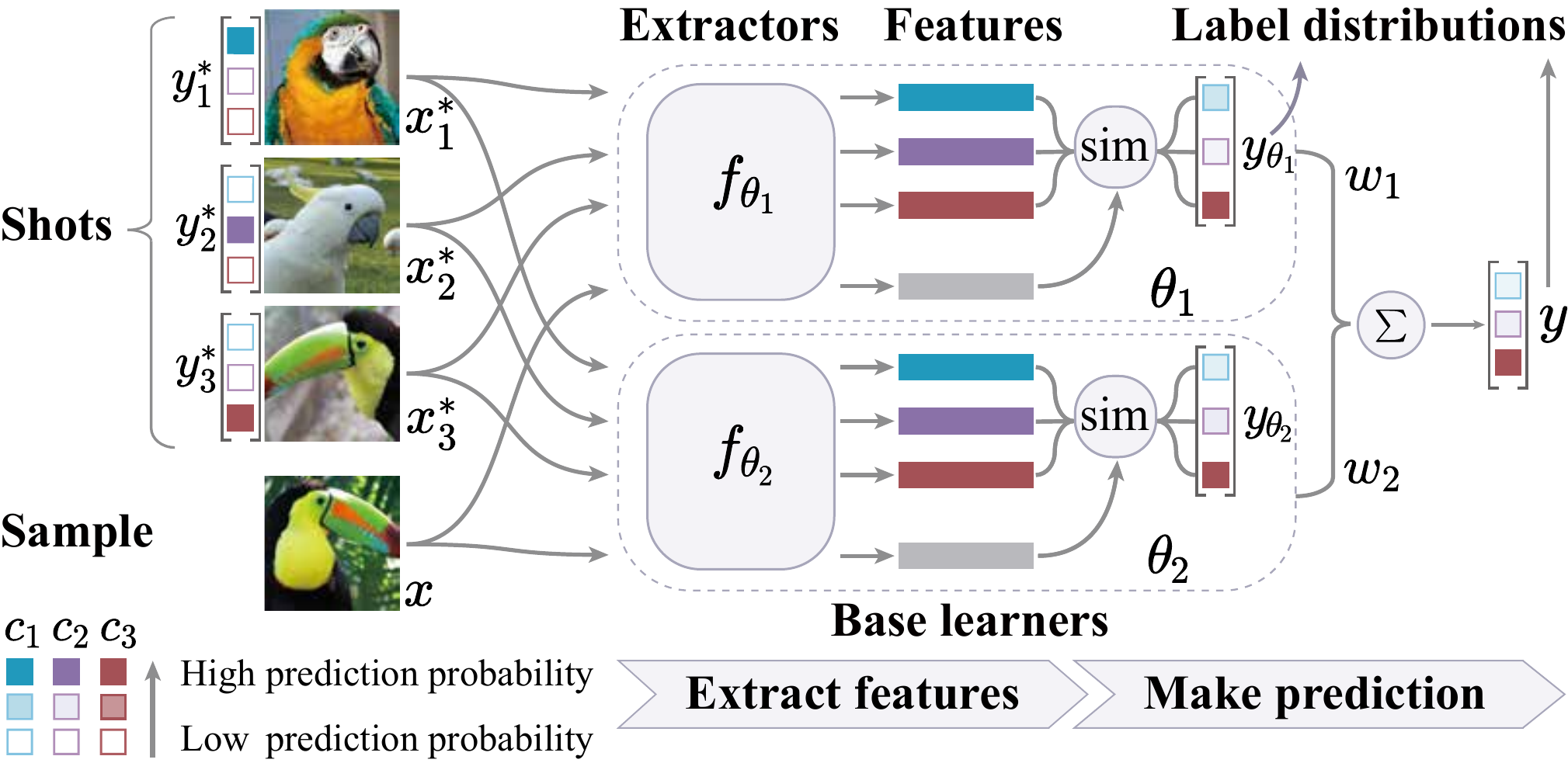}}
\put(-125,-4){(a)}
\put(-45,-4){(b)}
\caption{The prediction process of the ensemble few-shot classifier:
(a) each base learner extracts the features of the shots and samples;
(b) label distributions of the samples ($y_{\theta_1},y_{\theta_2}$) are calculated based on the similarity between the features and then averaged with weights to obtain the final label distribution $y$.
}
\label{fig:fsl_intro}
\end{figure}

\begin{figure*}[!t]
\centering
{\includegraphics[width=\linewidth]{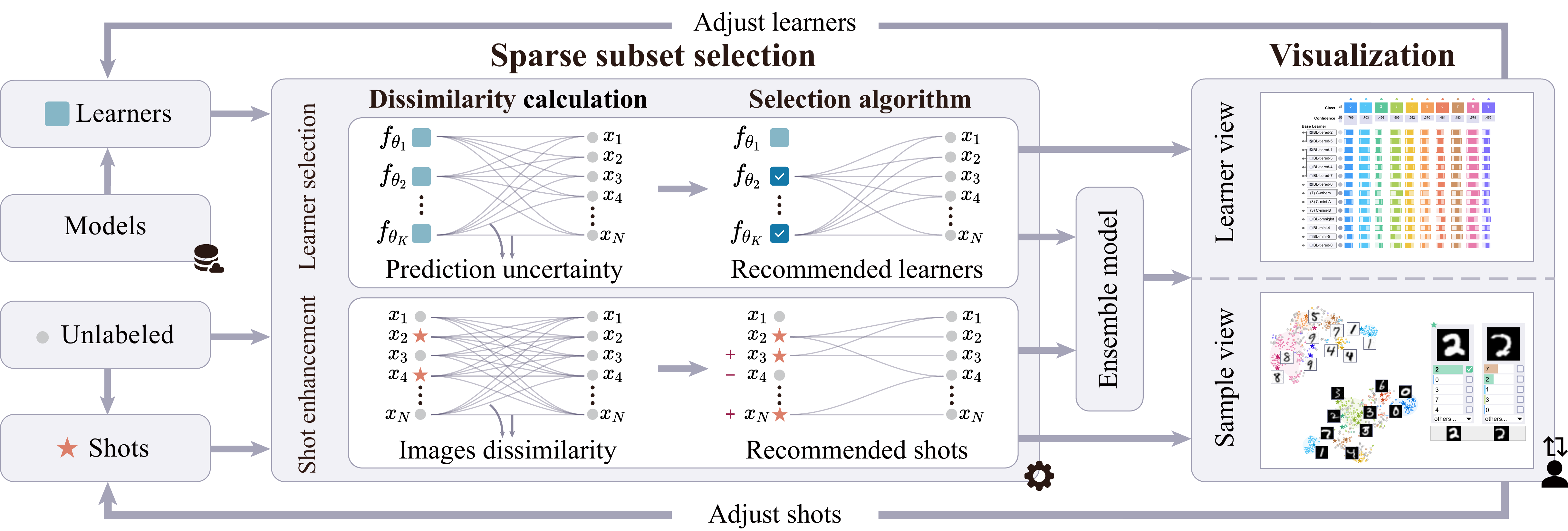}}
\caption{FSLDiagnotor overview. Given the base learners and samples with a few shots, the sparse subset selection module recommends base learners and shots for building the ensemble few-shot classifier.
The visualization module then explains how the learners and shots affect the final predictions, which facilitates users to improve them for interactively tuning the model.
}
\label{fig:pipeline}
\end{figure*}


Few-shot classification aims to learn a good classifier for unseen classes with a few shots.
Specifically, for the $N$ samples from these unseen classes, only the labels of $M$ shots (e.g., 1--5 shots per class) are provided. 
The shot set is denoted as $S=\{(x_j^\ast, y_j^\ast)\}_{j=1}^M$.
Here, $y_j^\ast$ is the label of shot $x_j^\ast$.
It is represented by a one-hot vector where the value of the corresponding class index is 1, and the others are 0s.
The goal is to build a model to predict the label distribution $y$ of a sample $x$ of the unseen classes based on $S$.
The label distribution $y$ is a $C$-dimensional probability vector. 
$C$ is the number of classes, and the value of the $i$-th dimension indicates the probability of the sample belonging to the $i$-th class.

Ensemble few-shot classification combines a set of base learners $\{\theta_k\}_{k=1}^K$ for achieving better performance.
\Fig{fig:fsl_intro} illustrates the process of predicting the label distribution of a sample based on three given shots and two learners.
For sample $x$, each learner $\theta_k$ generates a label distribution $y_{\theta_{k}}$. 
These label distributions are then averaged with weight $w_k$ to obtain the final label distribution $y$:\looseness=-1
\begin{equation}
y = \frac{1}{K} \sum_{k=1}^{K} w_k y_{\theta_{k}}.
\label{eq:ensemble_pred}
\end{equation}
$w_k$ is set to $1$ by default and can be adjusted in our tool.
It can be seen that the prediction results of the ensemble model are determined by the base learners and shots.
Thus users require a tool to help them examine the quality of base learners and shots and tune them for better performance.

%% file: 4.Requirements.tex
\section{Design of FSLDiagnotor}
\subsection{Requirement Analysis}
\label{sec:goals}

We collaborated with three machine learning experts (\E1, \E2, \E3) to design FSLDiagnotor.
\E1 is a postdoc researcher with an interest in data selection and few-shot learning.
\E2 and \E3 are two Ph.D. students with a focus on few-shot learning.
They are not the co-authors of this work.
The following three requirements are identified based on existing literature and three 60-minute participatory design sessions with the experts.

\textbf{R1: Tuning the selection of learners \liu{and their ensemble weights}}.
Previous work has indicated that the diversity and cooperation among the base learners are very important for improving the performance of the ensemble model~\cite{dvornik2019diversity}.
The experts also raised concerns regarding the current trial-and-error process for tuning the model when the accuracy is not acceptable.
They usually need to repeatedly examine the log data to understand the diversity and cooperation among learners, and manually adjust their selection and ensemble weights.
This is very time-consuming.
To facilitate the tuning process, the experts expressed the need to quickly understand the prediction behavior of base learners on different levels, including the overall difference compared with the ensemble model and the detailed difference on different classes.

\textbf{R2: Improving the quality of the shots}.
The representativeness of the shots is essential for few-shot classification~\cite{liu2018interactive}. 
As there are only a few labeled samples, mislabeled or confusing shots, such as the overlapped ones between two categories, decrease the model performance greatly.
Removing such low-quality shots and adding necessary new ones improve the coverage of the shots and overall performance.
When diagnosing an ensemble few-shot classifier, the experts need to understand the coverage of each shot and find the samples that are not well covered by the shots.
In addition, the experts required a tool that can automatically recommend low-quality shots to be removed and candidate samples to be added to the shot set, so that they can only examine a small subset and then quickly decide which ones to remove/add.

\textbf{R3: Being agnostic to the model architectures of learners}.
Existing methods for ensemble few-shot classification build the base learners based on a given model architecture~\cite{dvornik2019diversity,qi2020few}.
This is not flexible as a fixed model architecture cannot satisfy the performance requirements of different applications.
Thus, the experts need the flexibility to choose an appropriate architecture for a given task.
To directly employ different model architectures, such as a pre-trained ResNet model~\cite{He2016CVPR} or a newly developed few-shot learning model, the ensemble model should be agnostic to the model architectures that are used to build the base learners.

\subsection{System Overview}
Motivated by the requirements, we have developed FSLDiagnotor to interactively select high-quality base learners and shots.
As shown in \Fig{fig:pipeline}, it consists of two modules: sparse subset selection and visualization.
Given a set of base learners, shots, and unlabeled samples, the \textbf{sparse subset selection} module automatically recommends a subset of learners and a few shots.
With these recommendations, an ensemble few-shot classifier is built.
Next, the matrix visualization in the \textbf{visualization} module illustrates the performance of the learners and helps adjust their ensemble weights adaptively to improve the performance (\textbf{R1}).
Users can also examine the coverage of the shots in the scatterplot and replace the low-quality shots with the high-quality ones (\textbf{R2}).
The two modules work together to support an iterative tuning process until the desired performance is achieved.
During the process, users can directly adjust the selection of the base learners without considering their model architectures (\textbf{R3}).
This is achieved by building them directly on the features extracted by these models.
As such, the ensemble model focuses only on feature-level integration.
\liu{With this characteristic, users can directly use pre-trained models and newly developed few-shot models to extract features.
This} saves the training time and facilitates building the ensemble model flexibly.

%% file: 5.Algorithm.tex
\section{Sparse Subset Selection}
\label{sec:selection}

To build a high-quality few-shot classifier, FSLDiagnotor supports two tasks: 1) selecting a subset of \liu{diverse and cooperative} base learners; 2) enhancing the representativeness of shots by replacing the low-quality ones with the high-quality ones.
Because both tasks aim to find a small representative subset from a large data collection, we formulate them as \yang{distance-based} sparse subset selection~\cite{elhamifar2015dissimilarity}.
In this section, we first give an overview of the subset selection algorithm, then present how it can be extended to base learner selection and shot enhancement \yang{with task-related distances},
and finally give the time complexity analysis.
The quantitative result is shown in \Sec{sec:numerical_exp}.

\subsection{Algorithm Overview}
\label{sec:algorithm}
\begin{figure}[!t]
\centering
{\includegraphics[width=\linewidth]{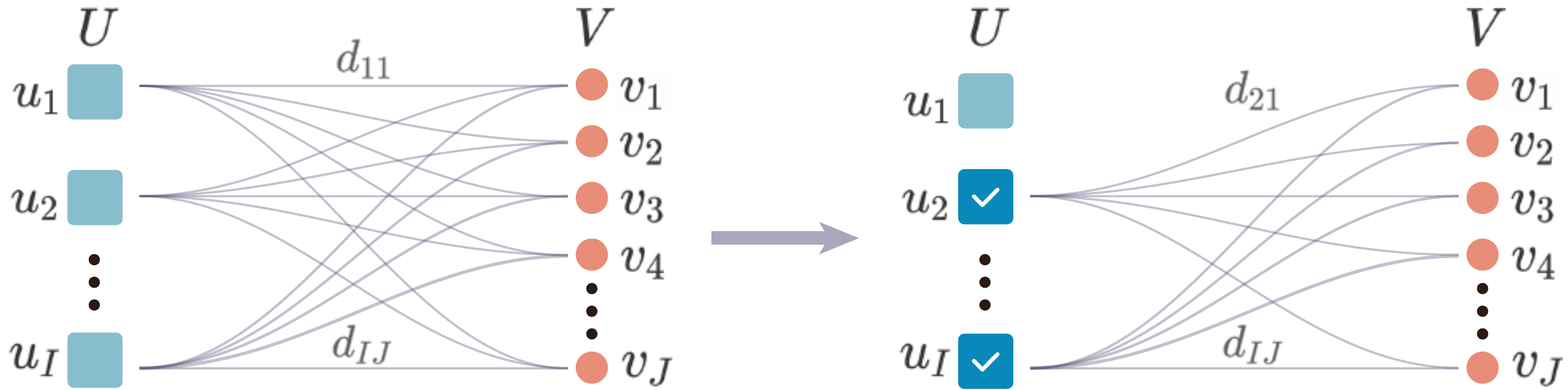}}
\put(-210,-14){(a) Input}
\put(-88,-14){(b) Selection result}
\caption{An example of sparse subset selection.}
\vspace{-3mm}
\label{fig:ds3}
\end{figure}

\Fig{fig:ds3} illustrates the basic idea of the algorithm.
Given two sets $U=\{u_i\}_{i=1}^{I}$ and $V=\{v_j\}_{j=1}^{J}$ ($U$ and $V$ can be identical or different), the sparse subset selection algorithm aims to find a subset of $U$ that can well represent set $V$.
This is achieved by minimizing the following function that balances the representation quality and the size of the subset:
\begin{equation}
\begin{aligned}
\sum_{j=1}^{J}&\sum_{i=1}^{I} z_{ij} d_{ij} + \alpha \sum_{i=1}^{I} \max_j z_{ij}
\\
\text{s.t.}\quad z_{ij}& \in\{0,1\},\ \forall i,j;\ \sum_{i=1}^{I} z_{ij}=1,\ \forall j.
\label{eq:cost_with_sparse_rewrite}
\end{aligned}
\end{equation}

The first term is the cost of representing $V$ with $U$ (representation cost), and the second term is the sparsity term to penalize a large subset.
In the first term, $z_{ij}$ is a binary variable indicating whether $v_j$ is represented by $u_i$, $d_{ij}$ is the distance between $u_i$ and $v_j$,
and the constraint $\sum_{i=1}^{I} z_{ij}=1$ guarantees that $v_j$ is represented by only an element in $U$.
In the second term, $\max_j z_{ij}=1$ if $u_i$ is selected in the subset, and $\sum_{i=1}^{I}\max_j z_{ij}$ is the size of the subset.
$\alpha\ge 0$ controls the trade-off between the two terms.

\yang{The proposed formulation is NP-hard~\cite{schrijver1998theory}.
To solve it efficiently,}
we relax the discrete 0-1 integer $z_{ij}\in\{0,1\}$ to $z_{ij}\ge 0$ and convert the sparse subset selection into a continuous optimization problem.
As in Elhamifar~\etal\cite{elhamifar2015dissimilarity}, we adopt the alternating direction method of multipliers framework to optimize \Eq{eq:cost_with_sparse_rewrite}.

\subsection{Base Learner Selection}
\label{sec:blselection}
Base learner selection aims to find a small subset of diverse and cooperative base learners to better predict the input samples (fitness).
Here, $U$ refers to the set of base learners $\{\theta_k\}_{k=1}^{K}$, and $V$ is the set of samples $\{x_i\}_{i=1}^{N}$.
As sparse subset selection encourages diversity among the selected learners, we then extend it by considering fitness and cooperation.
Accordingly, \Eq{eq:cost_with_sparse_rewrite} is rewritten as:
\begin{align}
\sum_{i=1}^{N}\sum_{k=1}^{K} z_{ki} d_{ki} &+\alpha_1\sum_{k=1}^{K}\lambda_k\max_i z_{ki}+ \alpha_2\sum_{1\le k<l\le K}\mu_{kl}\max_i z_{ki}\cdot \max_i z_{li}  \nonumber
\\
&\text{s.t.}\quad z_{ki} \ge 0,\ \forall k,i;\ \sum_{k=1}^{K} z_{ki}=1,\ \forall i,
\label{eq:learner_cost}
\end{align}
where the first term is the representation cost, the second term is the sparsity term that prefers the learners with higher fitness, and the third term is the cooperation term.
$\alpha_1$ and $\alpha_2$ control the trade-off among the three terms.
Following Elhamifar~\etal\cite{elhamifar2015dissimilarity}, $\alpha_1=\alpha_2=0.5\alpha_\text{max}$, $\alpha_\text{max}$ is the maximum distance between learners.\looseness =-1

In the first term, to calculate the representation cost, we need to define the distance between a base learner and a sample.
A straightforward way is based on the prediction accuracy.
However, we cannot evaluate the accuracy without ground-truth labels.
Instead, we use the prediction confidence to measure the distance \yang{because samples with high prediction confidence tend to be classified correctly}~\cite{zhu2009introduction}.
The prediction confidence of learner $\theta_k$ on $x_i$ is defined as the difference between the largest and the second-largest probabilities in the predicted label distribution $y_i$, which is denoted as $m_{ki}\in [0,1]$.
The distance between the learner $\theta_k$ and the sample $x_i$ is then defined by $d_{ki}=1-m_{ki}$ because we prefer the base learners with larger confidence $m_{ki}$.

In the second term, to encourage the selection of base learners with higher fitness,
we emphasize the ones that better predict the given shots.
A widely used measure, likelihood, is employed to estimate the fitness value.
Accordingly, we add $\lambda_k$ for each learner $\theta_k$, which is defined as its negative log-likelihood on the shots.

In the third term, to encourage the cooperation between two learners, $\theta_k$ and $\theta_l$, we penalize the difference 
\xia{between} their predictions.
\yang{Let $y_{ki}$ and $y_{li}$ be the label distribution of sample $x_i$ predicted by $\theta_k$ and $\theta_l$, respectively.}
Following the previous work of Dvornik~\etal\cite{dvornik2019diversity}, the prediction difference is measured by the symmetric KL-divergence between their predictions: \yang{$\mu_{kl}=\sum_{i=1}^{N}(\text{KL}(y_{ki}||y_{li})+\text{KL}(y_{li}||y_{ki}))/(2N)$.}
$\mu_{kl}$ is $0$ if the two learners make the same predictions.


\begin{figure*}[!tb]
\centering
{\includegraphics[width=\linewidth]{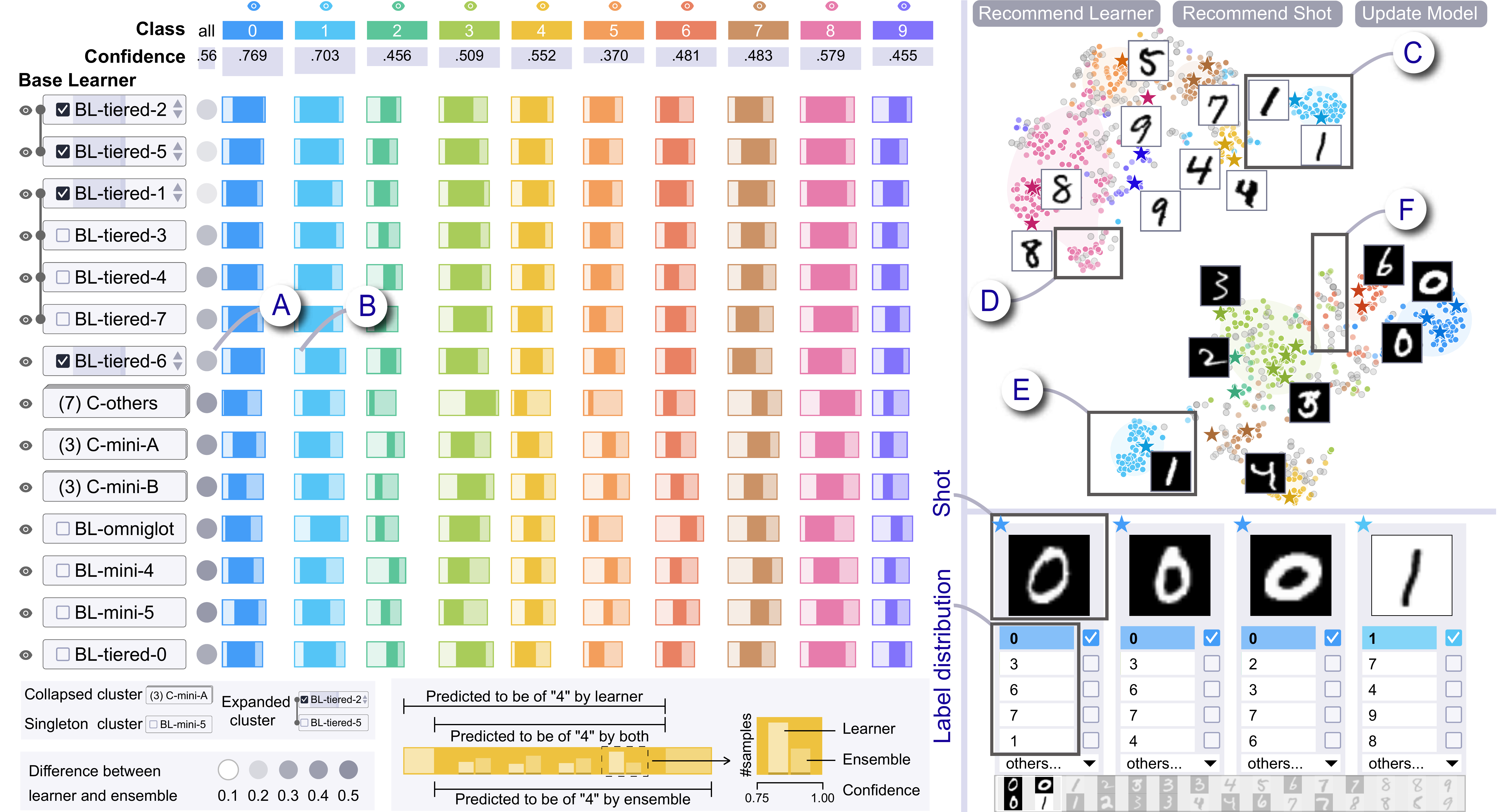}}
\put(-380,-9){\footnotesize (a) Learner view}
\put(-110,-9){\footnotesize (b) Sample view}
\vspace{-2mm}
\caption{FSLDiagnotor:
(a) \textit{learner view} compares base learners (rows) with the ensemble model, including the overall difference (circles in the first column) and detailed difference (stacked bars in the other columns);
(b) \textit{sample view} visualizes the shots and unlabeled samples in context.
The image content and label distributions of the samples of interest are displayed below.
}
\label{fig:teaser}
\end{figure*}

\subsection{Shot Selection}
\label{sec:shotselection}
Shot selection aims to find a very small set of shots that better represents all the samples.
Here, both $U$ and $V$ refer to the sample set $\{x_i\}_{i=1}^{N}$.
Rather than treating the samples equally in the sparsity term of \Eq{eq:cost_with_sparse_rewrite},
we tend to select the low-confidence samples with higher representativeness since selecting them as shots can help the model distinguish more low-confidence samples~\cite{yang2015multi}.
Moreover, we try to preserve the given shots to reduce the analysis burden and labeling efforts.
Accordingly, \Eq{eq:cost_with_sparse_rewrite} is rewritten as:

\begin{equation}
\begin{aligned}
&\sum_{j=1}^{N}\sum_{i=1}^{N} z_{ij} d_{ij} + \alpha\sum_{i=1}^N \beta_i\gamma_i\max_j z_{ij}
\\
\text{s.t.}&\quad z_{ij} \ge 0,\ \forall i,j;\ \sum_{i=1}^{N} z_{ij}=1,\ \forall j,
\label{eq:shot_cost}
\end{aligned}
\end{equation}
where the first term is the representation cost of the shots, and the second term is the sparsity term with preference on the previous shots.
\yang{$\alpha$ controls the number of recommended shots.
If we want to recommend $N_s$ shots, we then set $\alpha=\alpha_\text{max}/N_s$, where $\alpha_\text{max}$ is the maximum distance between samples.
}


In the first term, \liu{the distance between samples $x_i$ and $x_j$ is calculated by averaging the cosine distances between their features extracted by the selected base learners.}

In the second term, to encourage the selection of the low-confidence samples and given shots, we add a confidence coefficient $\gamma_i$ and a stability coefficient $\beta_i$ for $x_i$.
The confidence coefficient favors the selection of low-confidence samples with higher representativeness.
Accordingly, $\gamma_i$ is set to its average prediction confidence of the selected learners.
A sample with lower confidence results in a lower penalty in the sparsity term and then tends to be selected.
The stability coefficient aims to preserve the given high-quality shots.
Accordingly, $\beta_i$ is set to $0.1$ if $x_i$ is a given shot.
Otherwise, $\beta_i$ is set to $1$.
\subsection{Time Complexity Analysis}
\label{sec:time}
The time complexity of sparse subset selection is $O(|U||V|)$~\cite{elhamifar2015dissimilarity}.
As the number of learners is not large, the running time of the learner selection is usually acceptable.
However, the number of samples is relatively large,
and thus, the shot selection algorithm is rather slow in computation.
For example, it takes around 7 seconds to recommend shots from 1,000 samples.
To tackle this issue,
we first randomly sample a subset of samples and then recommend learners and shots based on the subset.
The effectiveness of this sampling strategy is evaluated in \Sec{sec:exp_sampling}.

%% file: 6.Visualization.tex
\section{FSLDiagnotor Visualization}
Although the sparse subset selection algorithm recommends a set of base learners and a few high-quality shots, the automatic recommendation results are not always perfect.
For example, using likelihood to measure the quality of base learners is sometimes not accurate since the number of shots is very limited.
In addition, 
\liu{an ambiguous shot wrongly representing some samples usually leads to more misclassification and thus the low representativeness of the shots.
}
To better explain the recommendation results and facilitate the interactive tuning of the recommended learners and shots, we design a visualization-based explanatory environment.
It consists of two components: 1) a {learner view} (\Fig{fig:teaser}(a)) to compare each base learner with the ensemble model in terms of prediction behavior (\textbf{R1}); and 2) a {sample view} (\Fig{fig:teaser}(b)) to present the shots and unlabeled samples in context (\textbf{R2}).
The two coordinated views enable users to easily adjust the shots and the learners without considering the architectures of the learners (\textbf{R3}).

\subsection{Learner View}
Due to the familiarity of users with the matrix visualization and its intuitiveness~\cite{Dinkla2012}, we employ it to compare a base learner with the ensemble model and different learners (\Fig{fig:teaser}(a)).
Users can tune the selection of learners or adjust their ensemble weights based on the comparative analysis.
\begin{figure}[!tb]
\centering
{\includegraphics[width=\linewidth]{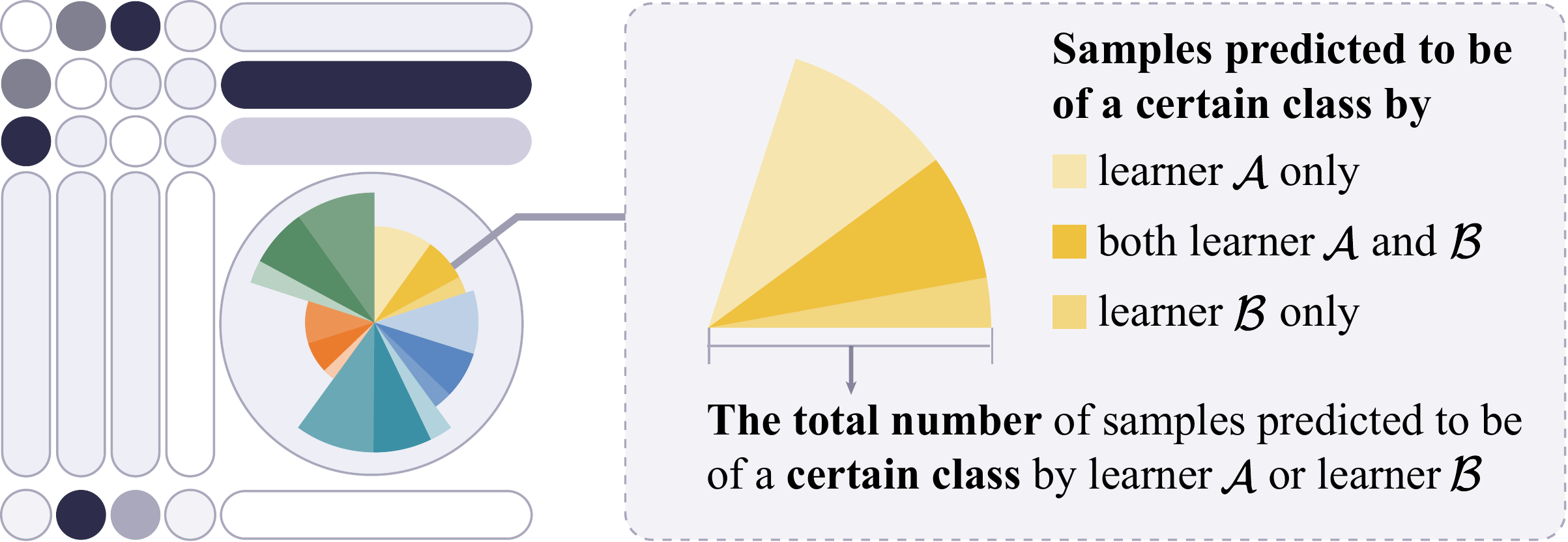}
\put(-255,-15){(a) Matrix with zoomable cells}
\put(-100,-15){(b) Encoding}
}
\caption{The alternative design of the {learner view}.
Both rows and columns represent base learners. A darker cell indicates a larger prediction difference between the two learners.
}
\vspace{-3mm}
\label{fig:old_design}
\end{figure}

\noindent\textbf{Visual design}.
Our first design focuses on the pairwise comparison between base learners, including the agreements and differences between the predictions of two learners.
We design a matrix with zoomable cells (\Fig{fig:old_design}(a)) to present the pairwise comparison results where rows and columns represent learners.
A sequential color scheme from white to black is used to encode the total number of samples that are predicted differently by the two learners.
Users can click on a cell of interest and zoom into it for the details of prediction behavior, which is depicted by a coxcomb chart.
In this chart, each sector represents a class that samples are predicted to be of.
A sector consists of three clockwise sub-sectors in the same hue (\Fig{fig:old_design}(b)), which represents the samples predicted to be of the same class by learner $\mathcal{A}$ only, by both learners $\mathcal{A}$ and $\mathcal{B}$, and by learner $\mathcal{B}$ only, respectively, where $\mathcal{A}$ is represented by the row, and $\mathcal{B}$ is represented by the column.
The total number of samples that are predicted to be of this class is encoded by the radius of the sector.
The experts agree that the comparison between two learners is helpful.
They like the design of three sub-sectors that illustrate the agreement and difference between two learners.
However, they are more interested in comparing a base learner with the ensemble model instead of comparing two learners (\textit{issue 1}).
The pairwise comparison fails to explain the role of a base learner in the ensemble predictions.
Another concern is that this design does not support the comparison across different base learners on a specific class (\textit{issue 2}), which is important for diagnosis.\looseness=-1

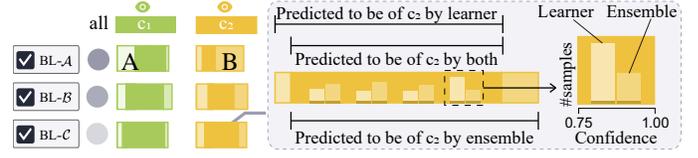
\begin{figure}[!tb]
\centering
{\includegraphics[width=\linewidth]{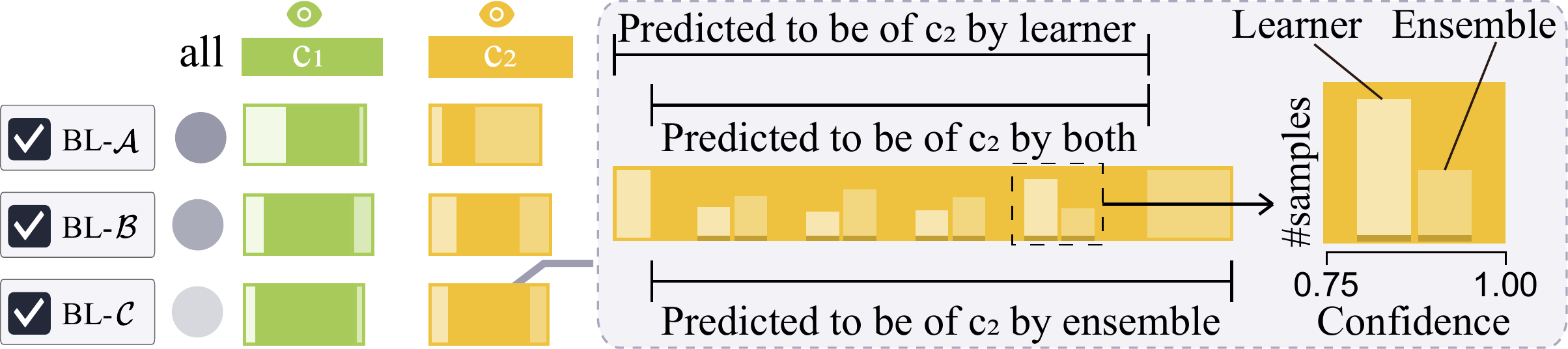}
\put(-255,-15){(a) Matrix with zoomable cells}
\put(-100,-15){(b) Encoding}
\put(-212,30){A}
\put(-174,30){B}
}
\caption{
The design of the {learner view}.
Rows represent base learners, columns represent classes, and cells disclose the agreements and differences between the predictions of the base learners and ensemble model.
\looseness=-1}
\vspace{-3mm}
\label{fig:new_design}
\end{figure}

To tackle these issues, we augment the matrix visualization to emphasize the comparison between the base learners and the ensemble model (\textit{issue 1}) and enable class-level comparison (\textit{issue 2}).
In the matrix visualization (\Fig{fig:new_design}(a)), each row represents a base learner.
The first column encodes the number of samples predicted differently between a base learner and the ensemble model (\textit{issue 1}) with a sequential color scheme.
The darker the cell is, the larger the difference is.
The remaining columns present the comparison between a base learner and the ensemble model (\textit{issue 1}) in terms of each class (\textit{issue 2}).
Instead of using the coxcomb chart in the first design, we employ a common visual metaphor, the stacked bar, to represent the agreement and difference between the predictions.
As shown in \Fig{fig:new_design}(b), the length of the stacked bar encodes the total number of the samples predicted to be of a certain class by the base learner and/or the ensemble model.
The hue of the stacked bar encodes the class.
As the experts are more familiar with the stacked bars, they can quickly identify the differences between each learner and the ensemble model under different classes.
\Fig{fig:new_design}A is an example where base learner ``BL-$\mathcal{A}$'' mostly agrees with the predictions made by the ensemble model on class ``$c_1$.''
However, there are many samples that are only recognized by ``BL-$\mathcal{A}$.''
As a result, the first bar is much longer than the third bar.
This indicates that ``BL-$\mathcal{A}$'' \textit{over-predicts} on class ``$c_1$.''
Similarly, we find that ``BL-$\mathcal{A}$'' \textit{under-predicts} on class ``$c_2$'' (\Fig{fig:new_design}B).

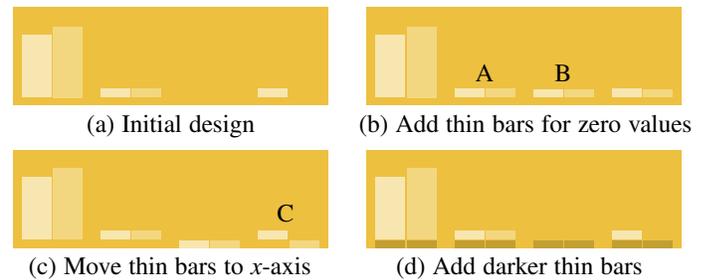
\begin{figure}[!b]
\centering
{
\vspace{-3mm}
\includegraphics[width=\linewidth]{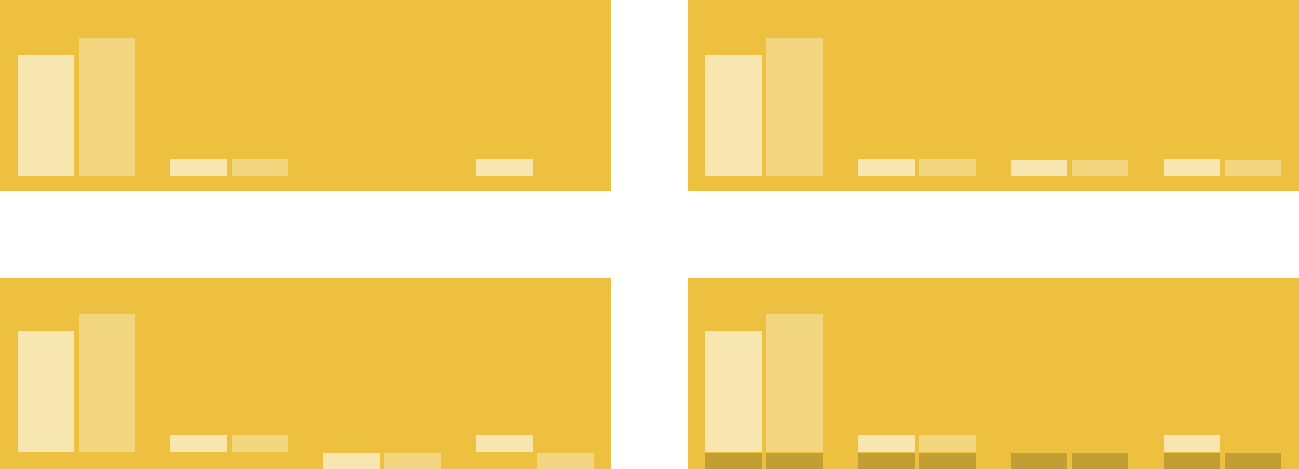}}
\put(-225,44){(a) Initial design}
\put(-122,44){(b) Add thin bars for zero values}
\put(-247,-10){(c) Move thin bars to $x$-axis}
\put(-108,-10){(d) Add darker thin bars}
\put(-78,63){A}
\put(-48,63){B}
\put(-153,10){C}
\caption{Four designs for comparing the prediction confidence of samples: (a)-(c) alternative designs; (d) our design.}
\label{fig:histogram}
\end{figure}

The experts give positive feedback to the new design during our interviews.
Later,
two experts express the need to investigate the prediction confidence of the samples.
After a thorough discussion, we use a histogram to convey the number of samples that are predicted by the learner/ensemble model with four different confidence bins (\Fig{fig:new_design}(b)).
However, if one bar is not shown in a confidence bin due to the zero value, it is inconvenient for users to identify which one is not displayed (\Fig{fig:histogram}(a)).
A straightforward solution is to preserve a minimum height for each bar (\Fig{fig:histogram}(b)).
However, such a thin bar (\Fig{fig:histogram}B) is difficult to be distinguished from other bars with very small values (\Fig{fig:histogram}A).
Another option is to place the thin bar on the $x$-axis to avoid such misunderstanding (\Fig{fig:histogram}(c)).
After using it, the experts point out that it may be misunderstood as a negative value (\Fig{fig:histogram}C).
To tackle this issue, we add a default thin darker bar for each item on the $x$-axis (\Fig{fig:histogram}(d)).\looseness=-1

\noindent \textbf{Visualization scalability}.
Although the matrix visualization helps users efficiently examine the predictions of learners, it suffers the scalability issue when the number of learners/classes increases.
To tackle this, we cluster similar learners (or classes) using agglomerative clustering~\cite{sneath1973numerical}.
The key of the clustering method is to calculate the distance between learners (or classes).
The distance between learners is measured by the symmetric KL-divergence of their predictions.
The distance between classes is calculated as the Euclidean distance in the feature space.
Since each class can be characterized by its shots, one common way to represent the class is by averaging the shot features (shot-based feature).
However, it can be inaccurate due to the scarcity of shots.
To compensate for this, we consider the word embedding of the class label (label-based feature), which is extracted by GloVe~\cite{pennington2014glove}, a widely used word embedding model.
We then obtain a more robust feature representation by concatenating the shot- and label-based features.
Several interactions are provided to explore the clusters.
For example, users can expand a cluster by double-clicking the associated rectangle and adjust the clustering result by dragging-and-dropping the rectangles.
The clusters of less interest can be hidden to minimize distraction by clicking $\vcenter{\hbox{\includegraphics[height=1.2\fontcharht\font`\B]{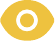}}}$.

\subsection{Sample View}
\begin{figure}[!b]
\centering
\vspace{-3mm}
{\includegraphics[width=\linewidth]{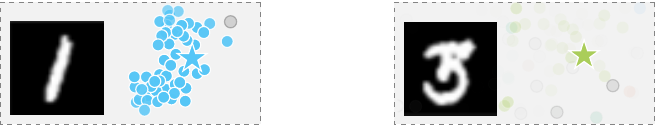}}
\put(-166,4){(a)}
\put(-15,4){(b)}
\caption{
The coverage of two shots: (a) a high-quality shot with many similar samples; (b) a low-quality shot with few similar samples.
}
\label{fig:coverage}
\end{figure}

\noindent\textbf{Visual design}.
The {sample view} (\Fig{fig:teaser}(b)) enables users to examine the shots in the context of samples and tune their selection.
\yang{For each sample, we first concatenate the features extracted by the base learners.} 
\liu{Next, to achieve better class separation~\cite{liu2016visualizing}, we employ t-SNE to project the samples onto 2D space and utilize a scatterplot to visualize the projections.
In the scatterplot, stars and circles are used} to represent shots and unlabeled samples, respectively.
Samples are colored according to their classes, and those with a confidence less than $0.2$ are colored \textit{gray}.
\yang{For each shot, we utilize a clutter-aware label-layout algorithm~\cite{meng2015clutter} to place the image content close to the shot and reduce the overlap with other scatter points.}
When users select the samples of interest, the image content and label distributions are displayed at the bottom of the view (\Fig{fig:teaser}(b)).
The label distributions are represented by colored bars, where the color encodes the class, and the length encodes the prediction probability.
Users can click the checkbox on the right side to add it as a shot or remove it from the shot set.\looseness=-1

The {sample view} also illustrates the influence of the base learners and shots in the ensemble model.
The influence of a learner is measured by the prediction confidence change of the ensemble model with/without the learner.
If the confidence of a sample increases by $0.2$ or more after adding the selected learner, the sample will be automatically marked with an upward arrow $\vcenter{\hbox{\includegraphics[height=1.6\fontcharht\font`\B]{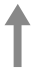}}}$.
If the confidence decreases by $0.2$ or more, the sample will be marked with a downward arrow $\vcenter{\hbox{\includegraphics[height=1.6\fontcharht\font`\B]{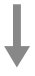}}}$.
We also use a gray density map \liu{as a guidance} to highlight the regions where a larger increase/drop in confidence happens (\Fig{fig:interaction}B).
Such regions indicate the conflicted predictions between the selected learner and ensemble model and need to be further checked.
The influence of a shot is characterized by its coverage, which contains its associated unlabeled samples with high similarity.
The associated unlabeled samples with higher similarity are encoded by darker class colors.
\Fig{fig:coverage} shows the coverage of two shots.
The first one is a high-quality shot of digit ``1'' since it influences a large number of neighboring samples that are correctly predicted with high confidence (\Fig{fig:coverage}(a)).
In contrast, the second one is a low-quality shot of digit ``3'' because it only covers a few samples predicted with low confidence (\Fig{fig:coverage}(b)).

\noindent\textbf{\yang{Visualization scalability}}.
\yang{The scatterplot inevitably suffers from the scalability issue when the number of samples grows~\cite{mayorga2013splatterplots}.
To address this issue, we build a hierarchy by utilizing the random sampling strategy in a bottom-up manner~\cite{xiang2019interactive}.
Random sampling is employed because it can well preserve the overall data distribution~\cite{yuan2020evaluation}.
When navigating the hierarchy, 
the sampled data at the current level are visualized using scatter points, and the others using a density map.
}

\subsection{Incremental Improvement of Learners/Shots}
\begin{figure}[!t]
\centering
{\includegraphics[width=\linewidth]{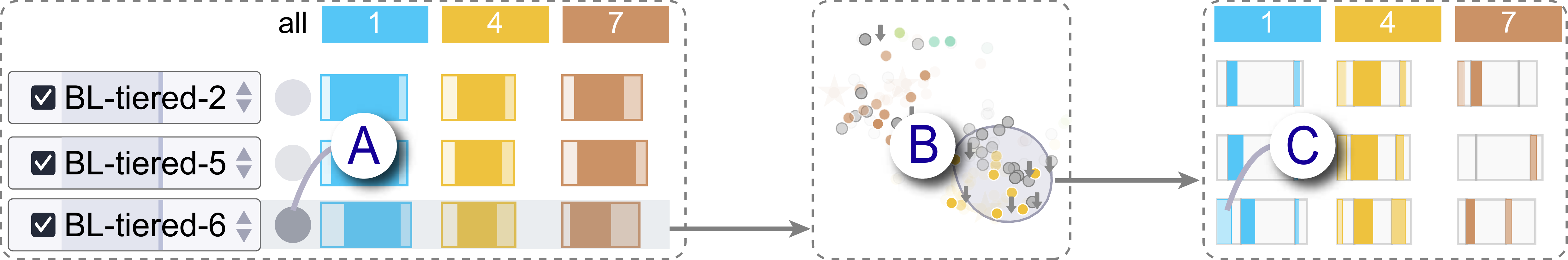}}
\caption{
``BL-tiered6'' causes a confidence drop in B, and C shows some samples in B are only predicted to be of ``1'' by ``BL-tiered6.''
}
\vspace{-3mm}
\label{fig:interaction}
\end{figure}
To facilitate the diagnosis of the ensemble few-shot classifier, FSLDiagnotor provides a few interactions to assist in 1) improving the selection of base learners; \liu{2) adjusting the ensemble weights of base learners; 3)} enhancing the quality of shots; \liu{4) mutually improving the learners and shots if either of them is adjusted.}
\yang{Here, recommendation-related interactions (e.g., recommend shots) and the weight adjustment are examples of semantic interactions~\cite{endert2016semantic}, which enable smooth communication between the user and the analytical model without direct manipulation of the model.}

\noindent\textbf{Improving the selection of base learners}.
FSLDiagnotor allows to remove low-quality learners and add high-quality ones.
To decide which one is of low/high quality, we allow users to 1) explore the influence of the learners on the ensemble model to identify the key samples that are predicted differently by them; and then 2) examine the prediction difference between the learners and the ensemble model on these samples.
For example, \Fig{fig:interaction} shows that there is a larger difference between ``BL-tiered6'' and the ensemble model (\Fig{fig:interaction}A).
Users can click ``BL-tiered6'' to examine its influence and find that it causes a large confidence drop in a region (\Fig{fig:interaction}B).
After selecting samples in this region using the lasso, these samples are highlighted on the associated bars with a solid filling style $\vcenter{\hbox{\includegraphics[height=1.6\fontcharht\font`\B]{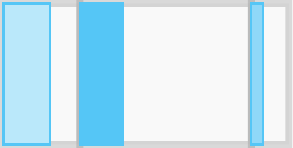}}}$.
From these bars, it can be seen that some samples are only predicted to be of ``1'' by ``BL-tiered6'' (\Fig{fig:interaction}C).
By clicking the associated bar (\Fig{fig:interaction}C), these samples are highlighted in the {sample view} for further examination.
If the selected learner makes many wrong predictions on these samples, users can remove it.\looseness=-1

\noindent\textbf{\liu{Adjusting the ensemble weights of base learners}}.
\liu{
The ensemble weight is important for the model performance. 
Although automatic weight adjustment is an efficient way to achieve this, it requires some extra validation samples with labels~\cite{zhou2012ensemble}. 
Since these validation samples are not available in few-shot applications,
FSLDiagnotor supports a semi-automatic adjustment of the ensemble weight of a learner to emphasize/de-emphasize it.}
\yang{For example, after examining a set of selected samples ($S_1$) that are predicted differently by the selected learner and the ensemble model, users can click $\blacktriangle$ to increase its weight if its predictions are mostly correct,
or click $\blacktriangledown$ to decrease its weight otherwise.
Since the exact weight is hard to decide, our tool automatically calculates the weight based on the prediction behavior of this learner and the ensemble model.
Specifically, $\forall x_j\in S_1$, the final prediction of the ensemble model $y_j$ should be consistent with $y'_j$, i.e.,  1) the prediction of the learner if users increase its weight or 2) the prediction of the ensemble model without the learner if users decrease its weight.
Moreover, $\forall x_j\in S_2$, where $S_2$ is the set of unselected samples, the final prediction $y_j$ should be as same as possible to the previous prediction $y_j^{\mathrm{prev}}$.
Based on the two considerations, the weight is decided by maximizing:}\looseness=-1
\begin{equation}
\begin{aligned}
\yang{\frac{\sum_{x_j\in S_1}\mathbb{I}(y_j=y'_j)}{|S_1|}+\frac{\sum_{x_j\in S_2}\mathbb{I}(y_j= y^{\mathrm{prev}}_j)}{|S_2|}}.
\label{eq:adjust_weight}
\end{aligned}
\end{equation}

\yang{The first term and the second term measure the prediction consistency on $S_1$ and $S_2$, respectively.
$\mathbb{I}(\cdot)$ is the indicator function. It equals 1 if the prediction is consistent, and 0 otherwise.
}



\noindent\textbf{\minor{Steering the selection of shots to enhance the quality}}.
FSLDiagnotor allows users to interactively enhance the quality of shots by removing the low-quality ones and adding necessary new ones \minor{in a steerable way}.
For example, 
users can identify the regions lacking shots and then label some of them.
As shown in \yang{\Fig{fig:BL-tiered6}B}, digits ``0''  (blue) are mostly misclassified to be of ``8'' (pink) since there are no shots of digit ``0'' in this region.
To improve the shot coverage in this region, users can manually add a few shots of ``0'' \liu{or click ``Recommend Shot'' to ask the tool to automatically recommend the candidate shots}. 
In addition, if one class is predicted with low confidence, users can examine the associated samples to figure out the potential reason.
Accordingly, users can click the bars in the matrix cell to examine the associated samples in the {sample view}.

\noindent\textbf{\liu{Mutually tuning the learners and shots}}.
\liu{In ensemble few-shot classification, learners and shots work together for the final predictions.
Generally, the change of learners influences the coverage of shots and vice versa.
Thus, if the learner set or the ensemble weights are changed, the shots should also be updated to adapt to the corresponding change. 
To this end, users click ``Recommend Shot.''
Then the shot selection algorithm is used to automatically recommend the shots.
On the other hand, if the shots are changed, users can click ``Recommend Learner'' to obtain a better combination of learners by the learner selection algorithm.
Such a process of mutual refinement saves users' time and efforts.}

%% file: 7.Evaluation.tex
\section{Evaluation}
We conducted \yang{three} experiments to evaluate the effectiveness of our subset selection algorithm.
We also demonstrated the usability of FSLDiagnotor through two case studies.
In the evaluation, we used the datasets with ground-truth labels \liu{to simulate the labeling process of users} and calculate the accuracy.  

\subsection{Quantitative Evaluation on Subset Selection}
\label{sec:numerical_exp}
\subsubsection{Datasets and Setups}

\noindent\textbf{Datasets}.
We evaluated the learner and shot selection algorithms with \xia{four widely used datasets: \textit{mini}-ImageNet~\cite{vinyals2016matching}, \textit{tiered}-ImageNet~\cite{ren2018meta}, MNIST~\cite{lecun1998gradient}, and CIFAR-FS~\cite{bertinetto2018metalearning}. 
}
\textit{Mini}-ImageNet consists of 80 seen classes and 20 unseen classes, each of which contains 600 images.
\textit{Tiered}-ImageNet contains 779,165 images of 608 classes (448 seen and 160 unseen classes).
The seen classes of these two datasets were used for training base learners, while the unseen classes were used to evaluate the performance of the model.
\xia{MNIST has 20,000 images of 10 unseen classes, and the images are augmented by inverting color.
CIFAR-FS has 12,000 images of 20 unseen classes.
}

\noindent\textbf{Base learners}.
We used 24 base learners in the ensemble model.
Sixteen of them are trained from scratch using ResNet-12 backbone~\cite{He2016CVPR}, where 8 of them are trained on different subsets of the seen classes of the \textit{mini}-ImageNet dataset, and the other 8 are trained on those of the \textit{tiered}-ImageNet dataset.
The remaining 8 base learners are pre-trained on external datasets, e.g., natural images in ImageNet~\cite{deng2009imagenet}, handwritten characters in Omniglot~\cite{lake2015human}.
We directly used the model parameters taken from publicly available implementations provided by Dvornik~\etal\cite{dvornik2020selecting}.

\noindent\textbf{Evaluation criteria}.
We evaluated the performance in terms of classification accuracy, which is averaged over 100 trials.\looseness=-1

\subsubsection{Effectiveness Evaluation of Sparse Subset Selection}
\label{sec:exp_effectiveness}
In this experiment, we evaluated whether the learner and shot selection algorithms can boost the few-shot classification accuracy on four datasets.
\yang{
Due to the limited number of shots, the randomness of few-shot classification is relatively high.
To reduce the effect of such randomness, more trials are needed~\cite{tian2020rethink}. 
To perform the evaluation efficiently, we used less samples for each class by following the common practice in few-shot learning~\cite{Boudiaf2020TIM}.
}
In particular, for \textit{mini}-ImageNet and \textit{tiered}-ImageNet, each task is a 5-class classification containing 5 randomly selected unseen classes, and each class contains 5 shots.
\yang{For MNIST and CIFAR-FS, we used all the unseen classes (10 and 20, respectively) in the tasks.
To simulate real-world applications, we do not guarantee that each class has the same number of shots.
Instead, we randomly select 30 and 60 samples as shots (each class has 3 shots on average) from these two datasets, respectively.}
Each class of the four datasets contains 15 unlabeled samples.
The \textbf{\liu{baseline}} is obtained by using all the base learners and initial random shots in the ensemble model.
Our method employs both the recommended base learners and recommended shots.
\yang{For a fair comparison, the number of the recommended shots is set to be the same as that of the initial shots.
The average number of the recommended learners over 100 trials is shown in Table~\ref{tab:result}.}
We compared our method, two ablations that only use either recommended learners (Rec. Learners) or shots (Rec. Shots), the state-of-the-art method, TIM~\cite{Boudiaf2020TIM}, and the baseline.

\begin{table}[!htb]
\centering
\caption{Classification accuracy on four datasets. \yang{The average numbers of recommended learners are given in parentheses.}}
\scalebox{0.85}{
\begin{tabular}{c|c|c|c|c}
\toprule
Model & \textit{mini} & \textit{tiered} & \yang{MNIST} & \yang{CIFAR-FS} \\
\hline

Baseline & 0.873 & 0.849 & \yang{0.476} & \yang{0.447} \\
TIM~\cite{Boudiaf2020TIM} & 0.874 & 0.898& \yang{-} & \yang{-}  \\
Rec. Shots & 0.877 & 0.862 & \yang{0.611}  & \yang{0.517} \\
Rec.\ Learners & 0.880 \yang{(3.9)}  & 0.868 \yang{(3.6)} & \yang{0.481} \yang{(4.3)}  & \yang{0.480} \yang{(5.2)}  \\
Our method &\textbf{0.896} \yang{(3.9)} & \textbf{0.908} \yang{(3.6)} & \textbf{\yang{0.615}} \yang{(4.3)} & \textbf{\yang{0.541}} \yang{(5.2)} \\
\bottomrule
\end{tabular}
}
\label{tab:result}
\end{table}

As shown in Table~\ref{tab:result}, using either recommended learners or shots alone can boost the performance on all datasets, and combining them together can further improve the performance.
\liu{By comparing the recommended learners/shots with the initial ones, we found that the low-quality learners/shots, such as a learner that has poor performance and predicts differently from the majority, were removed. 
Some high-quality learners/shots, such as a shot that well represents the unlabeled samples but does not appear in initial shots, were added.
This is the main reason why the developed subset selection algorithms can boost the performance.}

\subsubsection{Balance between Effectiveness and Efficiency}
\label{sec:exp_sampling}

\begin{table}[!b]
\centering
\caption{The accuracy using different sampling ratios (SR).}
\scalebox{0.71}{
\begin{tabular}{c|c|c|c|c|c|c|c|c|c}
\toprule
SR & 1\% & 2\% & 3\% & 4\% & 5\% & 6\% & 7\% & 10\%& 100\% \\
\hline
\textit{mini} & 0.838 & 0.865 & 0.874 & 0.882 & 0.886 & 0.889 & 0.891 & 0.891 & 0.893 \\
\textit{tiered} & 0.851 & 0.875 & 0.887 & 0.895 & 0.899 & 0.901 & 0.901 & 0.902 & 0.907 \\
\yang{MNIST}&\yang{0.591}&\yang{0.592}&\yang{0.597}&\yang{0.601}&\yang{0.598}&\yang{0.599}&\yang{0.603}&\yang{0.602}&\yang{0.609}\\
\yang{CIFAR-FS}&\yang{0.526}&\yang{0.531}&\yang{0.542}&\yang{0.548}&\yang{0.550}&\yang{0.551}&\yang{0.547}&\yang{0.554}&\yang{0.554}\\
\bottomrule
\end{tabular}}
\label{tab:performance_result}
\end{table}

\yang{
In our implementation, the random sampling strategy is employed to reduce the time cost for tasks with tens of thousands of samples or more.
}
Here, we conducted this experiment to 1) investigate whether this sampling strategy can reduce the time cost while achieving comparable performance to that of running the algorithm on all the samples, and 2) determine the smallest sampling ratio needed to meet this requirement.
\yang{
We adopted different sampling ratios (1\%, 2\%, 3\%, 4\%, 5\%, 6\%, 7\%, 10\%, 100\%) for recommending learners/shots, and calculated the accuracy on all the samples except shots.}

Table~\ref{tab:performance_result} shows that the accuracy increases with the number of samples when the sampling ratio is lower than $5\%$.
However, when the sampling ratio is greater than $5\%$, the pace of the increase begins to slow down.
Based on this observation, we drew the conclusion that using a small subset of samples can achieve comparable accuracy to that of using the full samples.
Furthermore, the sampling ratio of 5\% is a good balance between efficiency and accuracy.\looseness=-1

\subsubsection{\yang{Analysis on Diversity and Cooperation}}

\label{sec:diversity}

\yang{
The goal of this experiment is to evaluate the diversity and cooperation between learners.
We use the Jaccard Index to measure the diversity between learners, which is widely used to measure the difference between two sets~\cite{eck2009normalize}.
Let $S_{\theta_k}$ and $S_{\theta_l}$ be the set of high-confidence samples ($>0.2$) predicted by two learners $\theta_k$ and $\theta_l$, respectively.
The diversity is defined as $|S_{\theta_k}\cap S_{\theta_l}|/|S_{\theta_k}\cup S_{\theta_l}|$.
A smaller value indicates that the two learners are more diverse.
We use the symmetric KL-divergence to measure the cooperation between learners, which is introduced in \Sec{sec:blselection}.
A smaller value indicates that the two learners are more cooperative.
The diversity/cooperation of a set of learners is defined as the average of all pairwise diversity/cooperation between two learners.
Table~\ref{tab:diverse} shows that on all the datasets, our method recommends a set of more diverse and cooperative learners.
}

\begin{table}[!ht]
\setlength\tabcolsep{9.2pt}
\caption{\yang{Comparison of the diversity and cooperation between all learners and recommended learners. 
The smaller values indicate that the recommended learners are more diverse and cooperative.}
}
\scalebox{0.8}{
\begin{tabular}{lcccccc}
\toprule

\multicolumn{1}{c}{}      & \multicolumn{3}{c}{\yang{Diversity}} & \multicolumn{3}{c}{\yang{Cooperation}} \\
                          & \yang{All}      & \yang{Rec.} & \yang{Diff}    & \yang{All}      & \yang{Rec.}  & \yang{Diff}    \\
\hline
\yang{\textit{mini}}    & \yang{0.124} & \yang{0.063} & \yang{49.2\%} & \yang{0.959} & \yang{0.204} & \yang{78.7\%} \\
\yang{\textit{tiered}}  & \yang{0.138} & \yang{0.082} & \yang{40.6\%} & \yang{0.942} & \yang{0.336} & \yang{64.3\%} \\
\yang{MNIST}          & \yang{0.449} & \yang{0.208} & \yang{53.7\%} & \yang{1.045} & \yang{0.531} & \yang{49.2\%} \\
\yang{CIFAR-FS}        & \yang{0.321} & \yang{0.264} & \yang{17.8\%} & \yang{2.311} & \yang{0.961} & \yang{58.4\%} \\
\bottomrule
\end{tabular}}
\label{tab:diverse}
\end{table}

\vspace{-3mm}


\subsection{Case Studies}
\liu{In the case studies, we used the same 24 base learners employed in the quantitative evaluation.
To demonstrate the generalization of our approach to new tasks, we used the MNIST and CIFAR-FS datasets because there are no base learners pre-trained on them.
Based on the experiment results in Sec.~\ref{sec:numerical_exp}, we select a trial with higher accuracy for each dataset.
The experts started from the setting of recommending learners because it does not need any human involvement.}
When performing the case studies, we followed the pair analytics protocol~\cite{arias2011pair}, where the expert guided the exploration, and we interacted with the tool.
This protocol helps the experts focus more on the analysis of the model.

\subsubsection{MNIST Dataset}
In this case study, we collaborated with expert \E1 to understand and diagnose a model built on the MNIST Dataset~\cite{lecun1998gradient}.
She is interested in knowing how FSLDiagnotor supports the selection of base learners and the enhancement of the shots, thus improving the accuracy of the model.
The experiment in \Sec{sec:exp_sampling} indicates that a sampling ratio of 5\% can better balance performance and efficiency.
Thus, \E1 sampled $5\%\times 20,000=1,000$ samples.

\noindent\textbf{Overview}.
\E1 first observed that four base learners were recommended by FSLDiagnotor (\Fig{fig:teaser}(a)).
She then examined the selected base learners and noticed that ``BL-tiered6'' made many different predictions from the ensemble model (\Fig{fig:teaser}A).
This needed further investigation to figure out the reason.
In the {sample view}, she observed that the samples were separated into two groups, with the upper ones being samples of black digits with a white background (\eg, \Fig{fig:teaser}C, D), and the lower ones being samples of white digits with a black background (\eg, \Fig{fig:teaser}E, F).
Most of the regions were covered by the given shots well (\eg, \Fig{fig:teaser}C, E).
However, there were a few regions not covered by the shots (\Fig{fig:teaser}D, F), where some samples (in gray) were predicted with low confidence.
Using the 4 recommended base learners (\textbf{R1}), the accuracy was \textbf{0.513}.
The accuracy was calculated offline before and after the corresponding operations to verify the effectiveness of the improvement with our tool.

\begin{figure}[!t]
\centering
{\includegraphics[width=\linewidth]{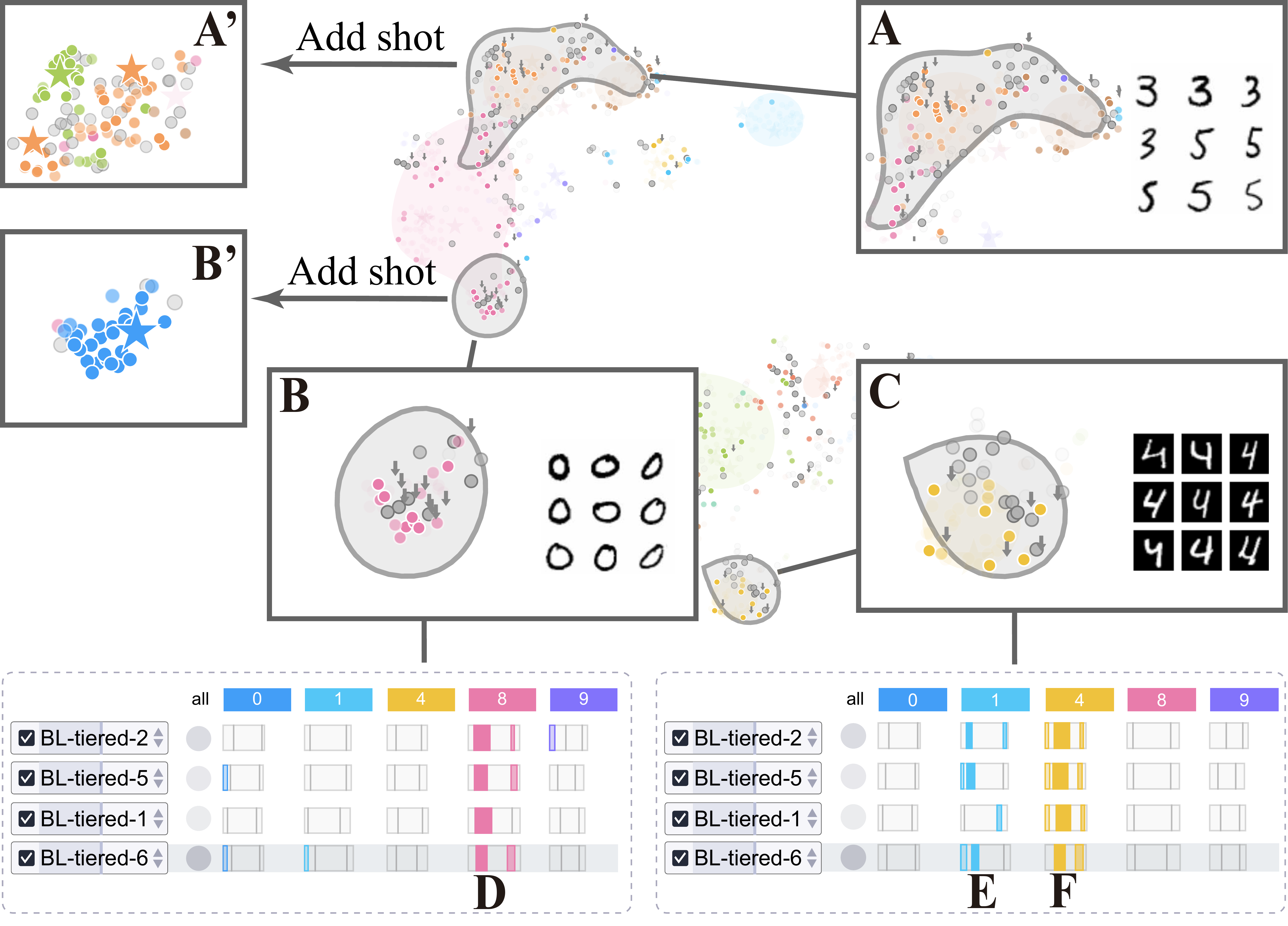}
}
\vspace{-5mm}
\caption{
Analyzing ``BL-tiered6.''
With it in the ensemble model, three regions (A, B, C) have a larger drop in prediction confidence.
}
\vspace{-3mm}
\label{fig:BL-tiered6}
\end{figure}

\noindent\textbf{Learner-based improvement}.
\E1 started the analysis from the base learners.
\E1 first examined the selected base learners for potential improvement.
Since ``BL-tiered6'' made more different predictions from the ensemble model,
she clicked on this learner to examine on which samples it made different predictions.
These samples were highlighted in the {sample view}.
Three gray density area also appeared in the {sample view}, indicating a larger drop in prediction confidence (\Fig{fig:BL-tiered6}A, B, C).
She decided to examine these three regions one by one.

\begin{figure}[!tb]
\centering
{\includegraphics[width=\linewidth]{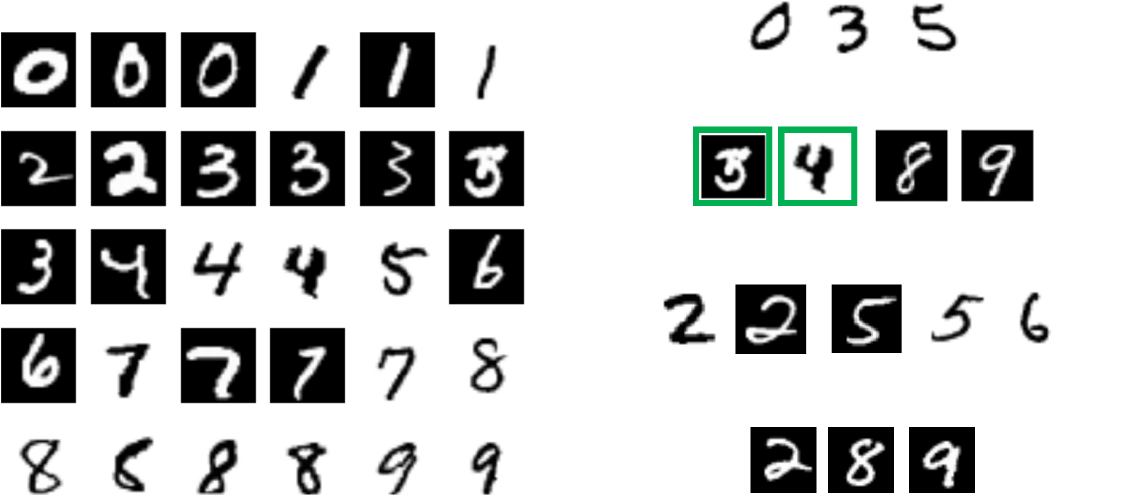}
\put(-220,-12){(a) Given shots}
\put(-101,90){(b) Add new shots}
\put(-113,56){(c) Enhance the given shots}
\put(-124,22){(d) Recommend shots in round 1}
\put(-124,-12){(e) Recommend shots in round 2}
}
\caption{Enhancing the quality of the given shots (a) by going through steps (b)-(e). The samples with green borders are recommended to be removed.}
\vspace{-3mm}
\label{fig:shot}
\end{figure}

\E1 began the analysis with \textbf{region A}, where most samples were black digits ``3'' and ``5.''
She noticed that the ensemble model misclassified most samples of ``3'' to be of ``5'' or ``8,'' and some samples of ``5'' to be of ``8.''
Checking the shots near this region, she found that there was no shot of digit ``3'' and only one shot of digit ``5.''
She decided to add more shots by selecting the samples in this region and clicking ``Recommend shot.''
Samples of ``3'' and ``5'' were recommended, which met her expectation.
She added one shot for ``3'' and one shot for ``5'' (\textbf{R2}).
Then \E1 switched to \textbf{region B}.
To her surprise, the region contained the samples of black digit ``0,'' but both the learners and the ensemble model misclassified them to be of ``8'' (\Fig{fig:BL-tiered6}D).
The reason was that there were no shots of black digit ``0'' (\Fig{fig:shot}(a)), so the nearest shots of a black digit ``8'' influenced the predictions of these samples.
These misclassifications can be corrected by adding more shots of black digit ``0.''
Since the samples of ``0'' in this region looked quite similar, she directly labeled one of them as a shot (\textbf{R2}).
\E1 further examined \textbf{region C}, where most samples were white digits ``4.''
The {learner view} showed that most of the base learners, as well as the ensemble model, made the correct predictions (\Fig{fig:BL-tiered6}F).
However, ``BL-tiered6'' misclassified some of them to be of ``1'' (\Fig{fig:BL-tiered6}E).
She noticed that ``BL-tiered6'' \textit{over-predicts} on ``1'' compared with other base learners (\Fig{fig:teaser}B).
She clicked the bar and found that many samples of white digits ``7'' were also predicted to be of ``1'' by ``BL-tiered6.''
She then concluded that ``BL-tiered6'' was confused about how to classify the white digits ``1,'' ``4,'' and ``7,'' which caused the drop in the prediction confidence (\Fig{fig:BL-tiered6}C).
Due to the poor performance of ``BL-tiered6'' in \textbf{region C}, she decided to remove it (\textbf{R1}).
After these adjustments, the model was updated.
The accuracy was improved from 0.513 to \textbf{0.582}.
\E1 was satisfied that 1) the added shots well covered those two regions (\Fig{fig:BL-tiered6}A', B'); 2) removing ``BL-tiered6'' increased the confidence of the samples in \textbf{region C} from 0.453 to 0.525.

\begin{figure}[!b]
\centering
{\includegraphics[width=\linewidth]{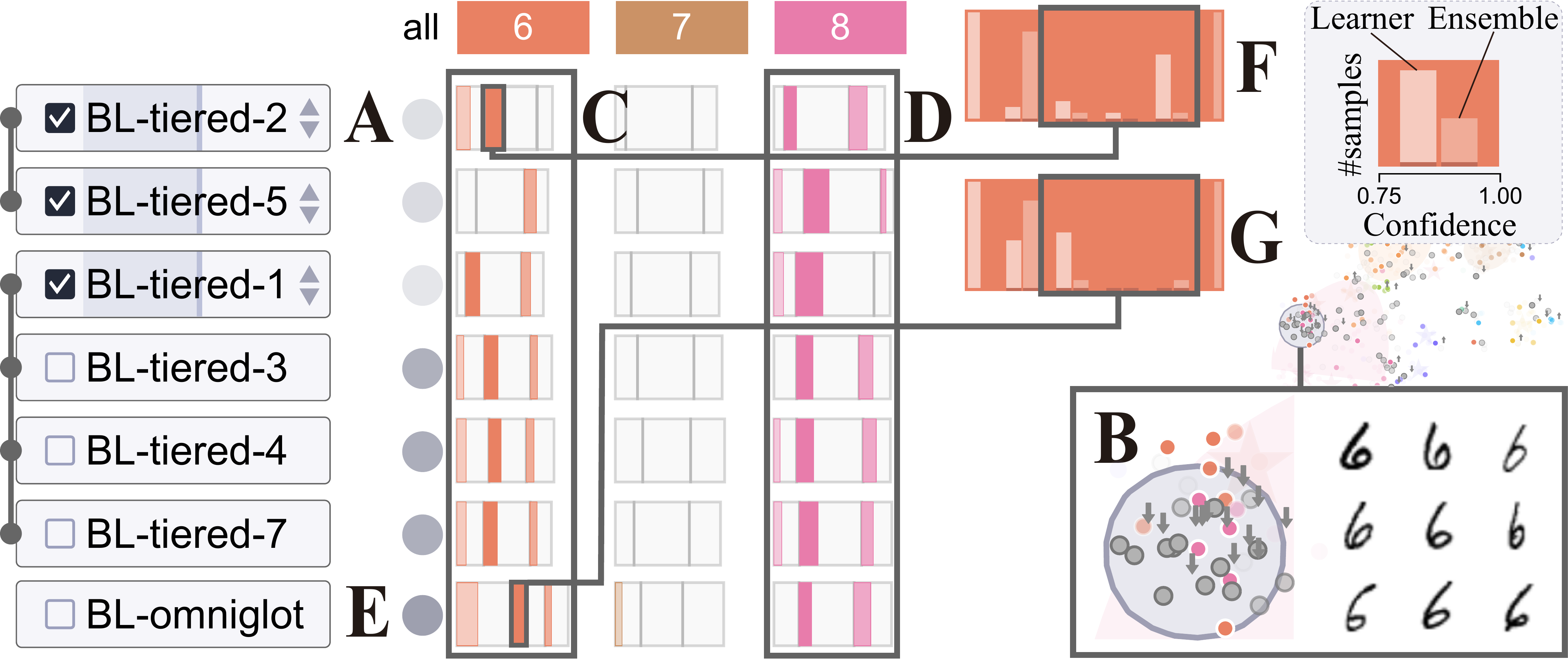}
}
\vspace{-3mm}
\caption{``BL-tiered2'' lowers the prediction confidence in region B.
These samples are of ``6'' but mis-predicted to be of ``8'' by the ensemble model.
In contrast, ``BL-tiered2'' and ``BL-omniglot'' make more correct predictions on them.}

\label{fig:omniglot}
\end{figure}

\noindent\textbf{Shot-based improvement}.
To adapt to the change of the base learners, she used our tool to automatically detect low-quality shots and recommend high-quality ones.
Inspired by the query strategy in active learning~\cite{settles2009active}, \E1 decided to add a few shots (3-5 shots) in each recommendation.
The recommended shots to be added/removed were displayed at the bottom of the {sample view}.
She found two low-quality shots with poor coverage (\eg, \Fig{fig:coverage}(b)) were detected and removed them (the samples in \Fig{fig:shot}(c) with a green border) by clicking the checkbox.
For the shots to be added, \E1 found that a white digit ``8'' and two white digits ``9'' were recommended, which did not appear in the given shots (\Fig{fig:shot}(a)).
To supplement the shot set that does not contain any white ``8'' and ``9,'' she selected one from each class, respectively (samples in \Fig{fig:shot}(c) without border).
\E1 updated the model with the new shot set, increasing the accuracy from 0.582 to \textbf{0.622}.
\E1 repeated the recommendation operation again and selected five more shots (\Fig{fig:shot}(d)), the accuracy was improved to \textbf{0.664}.

\noindent \textbf{Mutually tuning between the base learners and shots}.
To further improve the performance, she switched back to the {learner view} to see if there were any changes after updating the shots.
After examining the three selected learners one by one, she found that
``BL-tiered2'' (\Fig{fig:omniglot}A) lowered the prediction confidence of some samples in a small cluster.
This cluster contained some samples with black digits ``6'' (\Fig{fig:omniglot}B).
While ``BL-tiered2'' and ``BL-omniglot'' classified them correctly (\Fig{fig:omniglot}C), the ensemble model misclassified some of them to be of ``8'' (\Fig{fig:omniglot}D).
The distribution of confidence showed that these two learners were more confident than the ensemble model (\Fig{fig:omniglot}F, G).
As ``BL-tiered2'' was already selected in the ensemble model, \E1 \yang{clicked $\blacktriangle$ to increase} the weight of ``BL-tiered2.''
She also added ``BL-omniglot'' to the ensemble model.
She commented that Omniglot~\cite{lake2015human} was a dataset containing different handwritten characters and was similar to MNIST.
She considered that a learner trained on this dataset would be beneficial to the current task.
In addition, it increased the diversity among the learners as its predictions differed much from the ensemble model (\Fig{fig:omniglot}E).
After increasing the weight of ``BL-tiered2'' and adding ``BL-omniglot'' into the ensemble model (\textbf{R1}),
the accuracy was improved from 0.664 to \textbf{0.680}.
\E1 then added three more shots (\Fig{fig:shot}(e)) to adapt to the learner change (\textbf{R2}), and the accuracy increased to \textbf{0.707}.

\noindent\textbf{Summary}.
\E1 removed 2 low-quality shots and added 13 shots in total.
The final accuracy was $\textbf{0.707}$.
To achieve comparable performance, the random selection strategy requires 68 more shots and the automatic shot selection algorithm requires 28 more shots.
\E1 was satisfied with the ability of FSLDiagnotor in helping her identify misclassified regions and verify the recommended learners and shots for such a simple classification task.

\subsubsection{CIFAR-FS Dataset}

\begin{figure}[!b]
\centering
\vspace{-3mm}
{\includegraphics[width=\linewidth]{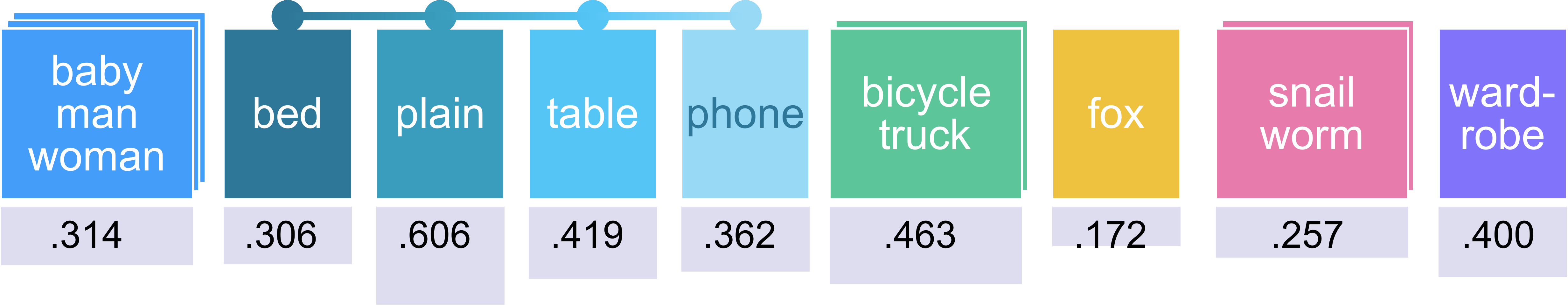}}
\vspace{-5mm}
\caption{The initial class clusters and their prediction confidence.}
\label{fig:class_cluster}
\end{figure}

This case study demonstrates the capability of our tool in boosting performance on a natural image dataset, CIFAR-FS~\cite{bertinetto2018metalearning}.
In this case study, we collaborated with \E2.
As the task involved more classes \yang{and contained only 12,000 samples}, \E2 increased the sampling ratio to $10\%$ and obtained $1,200$ samples.\looseness=-1

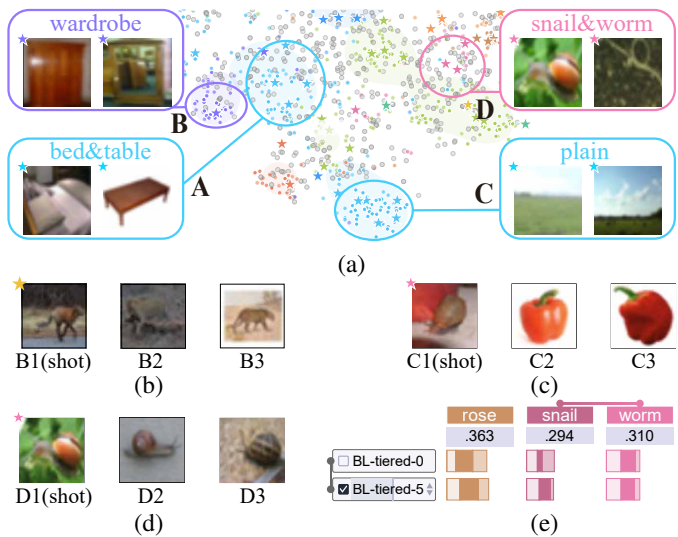
\begin{figure}[!tb]
\centering
{\includegraphics[width=\linewidth]{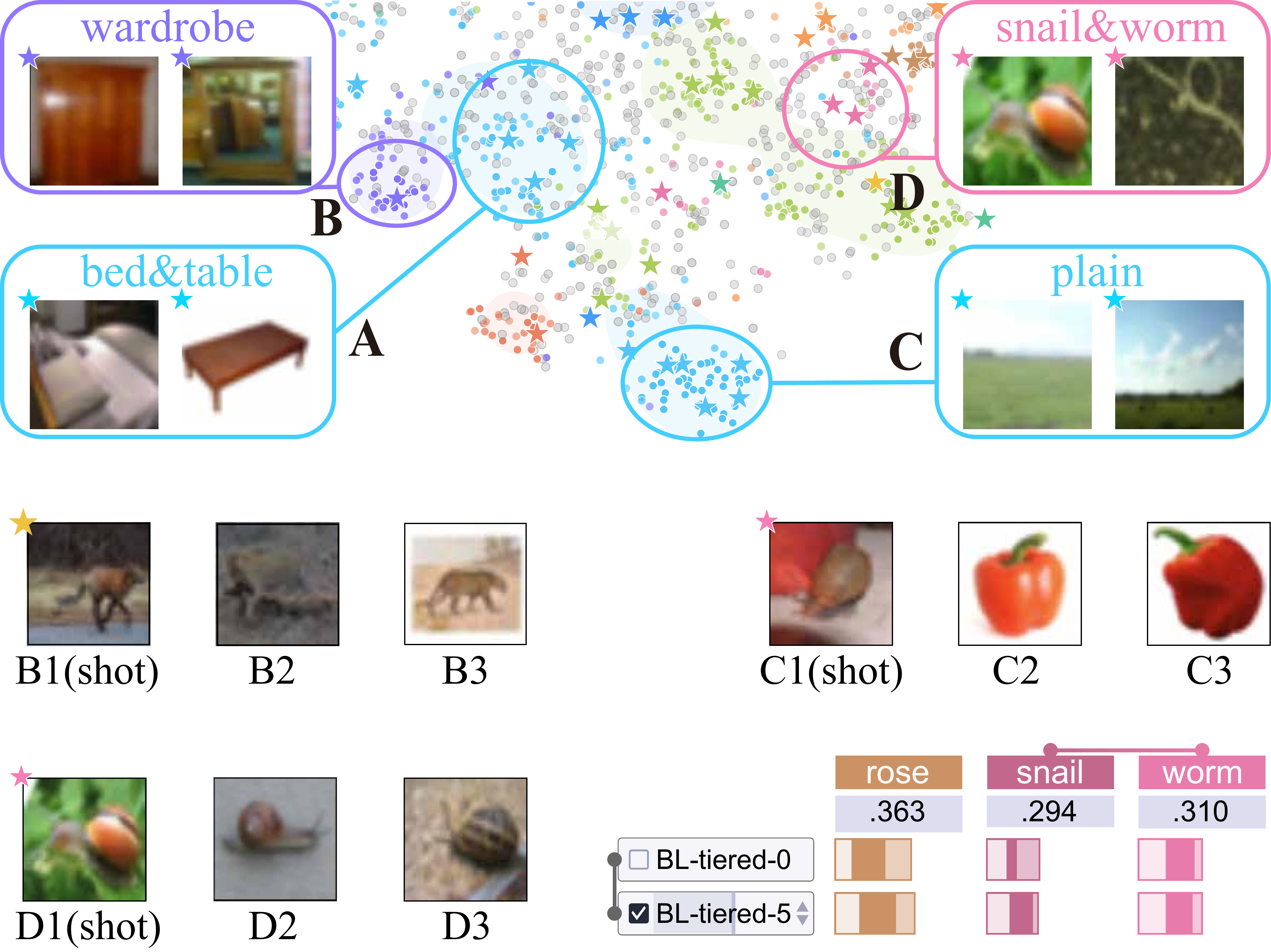}
\put(-128,90){(a)}
\put(-205,44){(b)}
\put(-55,44){(c)}
\put(-205,-8){(d)}
\put(-55,-8){(e)}
}
\vspace{-2mm}
\caption{
Analysis of the CIFAR-FS dataset:
(a) the {sample view};
(b) lack of shots for ``fox;''
(c) an outlier shot of ``snail;"
(d) poor diversity of the shots of ``snail;"
(e) the prediction behavior of two learners for ``snail."
}
\vspace{-3mm}
\label{fig:scatterplot}
\end{figure}

\noindent
\textbf{Overview}.
To improve readability, 20 classes were grouped into 10 clusters.
Most clusters looked reasonable.
For example, ``baby,'' ``man,'' ``woman'' were in the same cluster, and ``bicycle'' and ``truck'' were in another (\Fig{fig:class_cluster}).
However, \E2 found that ``plain,'' ``bed,'' ``table,'' ``phone'' formed a cluster while ``wardrobe'' formed another one.
The {sample view} (\Fig{fig:scatterplot}(a)) showed that ``bed'' (\Fig{fig:scatterplot}A), ``table'' (\Fig{fig:scatterplot}A), and ``wardrobe'' (\Fig{fig:scatterplot}B) were closed to each other, and ``plain'' (\Fig{fig:scatterplot}C) was away from them.
So he dragged ``wardrobe'' into the cluster and moved ``plain'' out as another cluster.
Seven base learners were recommended, including five ones trained on \textit{tiered}-ImageNet and two ones trained on \textit{mini}-ImageNet.
With the recommended learners, the accuracy of the ensemble was \textbf{0.497}.\looseness=-1

\noindent
\textbf{Diagnosing the clusters with poor performance}.
He first examined the cluster with the lowest confidence (0.172), which only contained the class ``fox'' (\Fig{fig:class_cluster}).
Twelve samples were predicted to be of ``fox.''
However, some of them were images with leopards (\Fig{fig:scatterplot}B2, B3).
In this class, there was only one shot (\Fig{fig:scatterplot}B1).
Both the learners and ensemble model have low confidence on the predictions.
\E2 commented that one shot was insufficient to distinguish ``fox'' from ``leopard.''
So he added 4 shots for ``fox'' and 2 shots for ``leopard'' (\textbf{R2}) and then updated the model.
The confidence increased to 0.288, and the accuracy reached \textbf{0.503}.\looseness=-1

Next, he moved to cluster ``snail\&worm'' with a lower confidence of 0.257.
After zooming in the cluster, he found that the confidence of ``snail'' was only 0.228, and many samples of ``pepper'' (\Fig{fig:scatterplot}C2, C3) were predicted to be of ``snail.''
\E2 examined the {sample view} and found a shot of ``snail'' that contained a red object  (\Fig{fig:scatterplot}C1).
He speculated that this shot disturbed the classification.
After removing it (\textbf{R2}), the confidence reached 0.294.
Then he examined the learner view and noticed the shorter length of the stacked bar charts for ``snail" (\Fig{fig:scatterplot}(e)).
This indicated that only a few samples were predicted to be of ``snail.''
To figure out why, he examined the base learners and found that ``BL-tiered5'' over-predicted on ``snail.''
The over-predicted samples were snails on non-green backgrounds (\Fig{fig:scatterplot}D2, D3) instead of the green background in the shots (\Fig{fig:scatterplot}D1).
He labeled three such samples to augment the diversity of the shots of ``snail'' (\textbf{R2}).
Moreover, he found that some samples of ``snail'' and ``worm'' were mixed and hard to be classified (\Fig{fig:scatterplot}D).
He used FSLDiagnotor to recommend two more shots for each of these two classes and updated the model (\textbf{R2}).
The confidence increased to 0.378, and the accuracy was \textbf{0.530}.

He continued to diagnose cluster ``baby\&man\&woman'' (confidence: 0.314) in a similar way, and the accuracy reached \textbf{0.541} after labeling 2 shots for ``man'' and 2 shots for ``woman'' (\textbf{R2}).
After removing the outlier shots in the largest cluster ``bed\&table\&phone\&wardrobe'' and adding six shots for ``bed'' and ``table'' (\textbf{R2}), the accuracy reached \textbf{0.561}.

\noindent
\textbf{Improving base learners and shots}.
The aforementioned diagnosis added/removed some shots.
To adapt to these changes, \E2 used FSLDiagnotor to recommend the learners and removed ``BL-mini7'' and ``BL-tiered5'' (\textbf{R1}).
The accuracy remained to be \textbf{0.561}.
To adapt to the change of the base learners, 14 more shots were also recommended (\textbf{R2}), and the accuracy increased to \textbf{0.594}.

\noindent\textbf{Summary}.
\E2 successfully improved the accuracy from 0.474 to 0.594 with only 37 extra shots.
To achieve comparable performance, the random selection strategy requires 115 extra shots, and the automatic shot selection algorithm requires 75 extra shots.
He was satisfied that FSLDiagnotor helped find a variety of quality issues of shots more efficiently.
``I do not realize that the shot like \Fig{fig:scatterplot}C1 hurts performance until I see those misclassified samples.''
He further pointed out that it was usually difficult to provide representative shots exhaustively.
The exploratory environment of the tool helps find the missing shots.

%% file: 8.Discussion.tex
\section{Expert Feedback and Discussion}
To evaluate the usefulness of FSLDiagnotor, we conducted six semi-structured interviews with the three collaborated experts (\E1, \E2, \E3) and three newly invited ones (\E4, \E5, \E6).
The three new experts are Ph.D. students who have worked in the field of machine learning for 5, 3, and 2 years, respectively.
\yang{In each interview, we spent 5 minutes introducing the design of our tool. 
Then the experts played with the tool to get familiar with it. 
For example, they tried to improve the performance by adjusting the selection of learners and/or enhancing the quality of shots.
Finally, we presented our case studies and gathered their feedback.}
Each interview lasted approximately 45-65 minutes.
All experts were generally positive about the usability of FSLDiagnotor.
They also pointed out a few limitations, which shed light on future work.


\subsection{Usability}

\noindent\textbf{\yang{Facilitating the performance improvement}}.
Encouragingly, our experts agreed that FSLDiagnotor was useful for improving model performance.
\E1 liked the shot quality enhancement module.
``Generally, some initial shots are probably of low-quality.
I would like to remove the low-quality ones and annotate a few more shots for better performance.
\yang{The tool recommends high-quality shot candidates for labeling, which reduces my workload.''}
\E5 was impressed by the promising accuracy improvement from 0.513 to 0.582 with only three shots added in the MNIST case.
\liu{In addition, the experts indicated that FSLDiagnotor not only provided an effective way to address the scenarios where only a few shots were available, but also an efficient mechanism to label a set of diverse shots that can better represent the unlabeled samples.}

\noindent\textbf{\liu{Being easy to use and reducing analysis efforts}}.
The experts agreed that the visual design was familiar and easy to understand.
\E2 commented, ``The stacked bar chart is very intuitive and clearly explains the prediction agreement and difference between the learner and ensemble model.''
\E4 believed that our tool could be used by practitioners easily,
``They are familiar with bar charts and scatterplots, so according to my experience, 15-30 minutes should be enough for them to get familiar with this tool.''
\yang{\E3 shared his experience of improving the performance, 
``An effective way to improve the performance is to label more shots in the regions that contain many samples with low confidence.
Such regions are highlighted with gray density and easy to identify.
With the recommendation function, I only examine the recommended samples from these regions and label the appropriate ones.''
}
\liu{The experts also commented that although extra analysis of the learners and samples was needed, the efforts were small because of the visual guidance and semantic interactions. 
Thus, their overall analysis efforts were reduced.} 

\subsection{Limitations}

\noindent\textbf{Generalization}.
In addition to classification tasks, the experts also expressed the need to apply our tool to handle object detection and segmentation.
After discussion, we found that the only change was induced by the IoU (Intersection over Union) measure employed in these tasks, which represents the area ratio of the intersection to the union of two shapes.
Unlike the binary variable used in the classification task to indicate whether a sample belongs to a class or not, the IoU score is a value between 0 and 1.
An interesting problem worth studying is how to effectively convey the distribution of the IoU scores in the {learner view}.
In addition, the experts expressed their need to analyze non-ensemble few-shot models, such as generative models and meta-learning ~\cite{wang2020generalizing}.
The subset selection algorithm and the {sample view} could be directly used to enhance the quality of shots and make adjustments.
However, the {learner view} needs to be re-designed to adapt to the analysis of a single model. We leave this as future work.

\noindent\textbf{Algorithm scalability}.
\liu{The shot recommendation is frequently performed to improve the performance in the analysis process. 
The experts usually select a region for the detailed examination, which contains at most thousands of samples.
For such cases, the subset selection algorithm can recommend shots in real-time.
However, when first providing the overview in the pre-processing stage, the recommendations are from the whole dataset.} 
It may still introduce the scalability issue \liu{into this offline process} when the dataset consists of tens of thousands of samples or more.
For example, it takes around $20$ hours to recommend shots from $100,000$ samples.
It is worth studying how to reduce the pre-processing time. \minor{For example, we can study how to use progressive visual analytics techniques~\cite{fekete:hal-01361430,fekete:hal-01202901} to recommend necessary shots progressively.}

\subsection{Lesson Learned}

\noindent\yang{\textbf{Using simple and familiar visualization}.
During the interviews, the experts appreciated the simple and familiar visual designs used in our tool.
A simple and intuitive visualization requires less time to learn and allows them to focus more on their analysis tasks. 
For example, \E2 commented,
\yang{
``The {learner view} can be regarded as a variant of the confusion matrix, with which I am very familiar.
Thus, I can go directly to analyze the root cause of low performance, which saves my time and efforts."  
}
\liu{The experts also pointed out that the visualization could be used in other tasks. For example, all the experts commented that they would like to use this tool to analyze a generic ensemble model, which they commonly used in various tasks.
The experts also indicated that the learner view can be directly used to compare datasets from different perspectives.
}
}

\noindent\minor{\textbf{Employing steerable visualization}.
During the development of FSLDiagnotor, we find that steerable visualization is an effective method to address the scalability issue when handling large-scale data.
The core of steerable visualization is to steer the computational efforts to the regions of interest~\cite{williams2004steerable}.
In FSLDiagnotor, since recommending shots may take a long time, users first identify the regions that lack shots and then steer more computational efforts to recommend shots in those regions.
Such steerable shot selection supports the exploration tasks where only a small subset of samples are of interest, such as finding diverse and well-performing learners from a large collection to build an ensemble model.
}

\noindent\liu{\textbf{Providing semantic interactions}.
In many sensemaking processes, users need to adjust the adopted analytical model to form hypotheses and derive conclusions.
Most existing interaction techniques rely on users’ expertise to adjust analytical models, such as modifying parameters and adding constraints. 
This requires users to be familiar with the working mechanism of the analytical model and thus limits the usage of the developed visual analysis tool/method.
With semantic interactions, users can easily steer the model without expertise in it.
Traditional interactions are well studied and several taxonomies are built~\cite{yi2007toward,heer2012interactive}.
However, semantic interaction research is quite new and more work is needed to form a taxonomy.
In FSLDiagnotor, we provide a few concrete examples of semantic interactions.
We hope \yang{these examples} can help inspire more research in this direction and build a solid taxonomy for semantic interactions.
}

%% file: 9.Conclusion.tex
 \section{Conclusion}
We have presented a visual analysis tool, FSLDiagnotor, to assist in visually diagnosing an ensemble few-shot classifier for better performance. 
FSLDiagnotor integrates the sparse subset selection method with an enhanced matrix visualization and a scatterplot to understand the inner workings of the base learners and the coverage of the shots.
With such a comprehensive understanding, users can build a better ensemble few-shot learning model by interactively and efficiently improving the selection of base learners and shots.
A quantitative evaluation demonstrates the effectiveness of the developed subset selection method in selecting appropriate base learners and enhancing the quality of the shots.
Two case studies are conducted to demonstrate the usefulness of our tool in diagnosing the few-shot classifier and improving its performance.

%% file: main.bbl
\begin{thebibliography}{10}
\providecommand{\url}[1]{#1}
\csname url@samestyle\endcsname
\providecommand{\newblock}{\relax}
\providecommand{\bibinfo}[2]{#2}
\providecommand{\BIBentrySTDinterwordspacing}{\spaceskip=0pt\relax}
\providecommand{\BIBentryALTinterwordstretchfactor}{4}
\providecommand{\BIBentryALTinterwordspacing}{\spaceskip=\fontdimen2\font plus
\BIBentryALTinterwordstretchfactor\fontdimen3\font minus
  \fontdimen4\font\relax}
\providecommand{\BIBforeignlanguage}[2]{{%
\expandafter\ifx\csname l@#1\endcsname\relax
\typeout{** WARNING: IEEEtran.bst: No hyphenation pattern has been}%
\typeout{** loaded for the language `#1'. Using the pattern for}%
\typeout{** the default language instead.}%
\else
\language=\csname l@#1\endcsname
\fi
#2}}
\providecommand{\BIBdecl}{\relax}
\BIBdecl

\bibitem{dvornik2019diversity}
N.~Dvornik, C.~Schmid, and J.~Mairal, ``Diversity with cooperation: Ensemble
  methods for few-shot classification,'' in \emph{Proceedings of the
  International Conference on Computer Vision}, 2019, pp. 3723--3731.

\bibitem{wang2020generalizing}
Y.~Wang, Q.~Yao, J.~T. Kwok, and L.~M. Ni, ``Generalizing from a few examples:
  A survey on few-shot learning,'' \emph{ACM Computing Surveys}, vol.~53,
  no.~3, pp. 1--34, 2020.

\bibitem{cvpr_challenge}
``Cross-domain few-shot learning ({CD-FSL}) challenge,''
  \url{https://www.learning-with-limited-labels.com}, 2019, {L}ast accessed
  2022-6-5.

\bibitem{kaggle_competition}
``Humpback whale identification,''
  \url{https://www.kaggle.com/c/humpback-whale-identification/overview}, 2019,
  {L}ast accessed 2022-6-5.

\bibitem{liu2018interactive}
S.~Liu, C.~Chen, Y.~Lu, F.~Ouyang, and B.~Wang, ``An interactive method to
  improve crowdsourced annotations,'' \emph{IEEE Transactions on Visualization
  and Computer Graphics}, vol.~25, no.~1, pp. 235--245, 2019.

\bibitem{qi2020few}
M.~Qi, J.~Qin, X.~Zhen, D.~Huang, Y.~Yang, and J.~Luo, ``Few-shot ensemble
  learning for video classification with slowfast memory networks,'' in
  \emph{Proceedings of the ACM International Conference on Multimedia}, 2020,
  pp. 3007--3015.

\bibitem{renggli2021a}
C.~Renggli, L.~Rimanic, N.~M. Gurel, B.~Karlas, W.~Wu, and C.~Zhang, ``A data
  quality-driven view of {MLOps},'' \emph{IEEE Data Engineering Bulletin},
  vol.~44, no.~1, pp. 11--23, 2021.

\bibitem{ng2021mlops}
A.~Ng, ``{MLOps}: From model-centric to data-centric {AI},'' 2021.

\bibitem{keim2008visual}
D.~A. Keim, F.~Mansmann, J.~Schneidewind, J.~Thomas, and H.~Ziegler, ``Visual
  analytics: Scope and challenges,'' in \emph{Visual Data Mining}.\hskip 1em
  plus 0.5em minus 0.4em\relax Springer, 2008, pp. 76--90.

\bibitem{tian2020rethink}
Y.~Tian, Y.~Wang, D.~Krishnan, J.~B. Tenenbaum, and P.~Isola, ``Rethinking
  few-shot image classification: A good embedding is all you need?'' in
  \emph{Proceedings of the European Conference on Computer Vision}, 2020, pp.
  266--282.

\bibitem{dvornik2020selecting}
N.~Dvornik, C.~Schmid, and J.~Mairal, ``Selecting relevant features from a
  multi-domain representation for few-shot classification,'' in
  \emph{Proceedings of the European Conference on Computer Vision}, 2020, pp.
  769--786.

\bibitem{He2016CVPR}
K.~He, X.~Zhang, S.~Ren, and J.~Sun, ``Deep residual learning for image
  recognition,'' in \emph{Proceedings of the IEEE Conference on Computer Vision
  and Pattern Recognition}, 2016, pp. 770--778.

\bibitem{hohman2018visual}
F.~Hohman, M.~Kahng, R.~Pienta, and D.~H. Chau, ``Visual analytics in deep
  learning: An interrogative survey for the next frontiers,'' \emph{IEEE
  Transactions on Visualization and Computer Graphics}, vol.~25, no.~8, pp.
  2674--2693, 2019.

\bibitem{yuan2021survey}
J.~Yuan, C.~Chen, W.~Yang, M.~Liu, J.~Xia, and S.~Liu, ``A survey of visual
  analytics techniques for machine learning,'' \emph{Computational Visual
  Media}, vol.~7, no.~1, pp. 3--36, 2021.

\bibitem{liu2016towards}
M.~Liu, J.~Shi, Z.~Li, C.~Li, J.~Zhu, and S.~Liu, ``Towards better analysis of
  deep convolutional neural networks,'' \emph{IEEE Transactions on
  Visualization and Computer Graphics}, vol.~23, no.~1, pp. 91--100, 2017.

\bibitem{bilal2017convolutional}
A.~Bilal, A.~Jourabloo, M.~Ye, X.~Liu, and L.~Ren, ``Do convolutional neural
  networks learn class hierarchy?'' \emph{IEEE Transactions on Visualization
  and Computer Graphics}, vol.~24, no.~1, pp. 152--162, 2018.

\bibitem{kahng2017activis}
M.~Kahng, P.~Y. Andrews, A.~Kalro, and D.~H. Chau, ``{ActiVis}: Visual
  exploration of industry-scale deep neural network models,'' \emph{IEEE
  Transactions on Visualization and Computer Graphics}, vol.~24, no.~1, pp.
  88--97, 2018.

\bibitem{liu2017analyzing}
M.~Liu, J.~Shi, K.~Cao, J.~Zhu, and S.~Liu, ``Analyzing the training processes
  of deep generative models,'' \emph{IEEE Transactions on Visualization and
  Computer Graphics}, vol.~24, no.~1, pp. 77--87, 2018.

\bibitem{wang2018dqnviz}
J.~Wang, L.~Gou, H.-W. Shen, and H.~Yang, ``{DQNV}iz: A visual analytics
  approach to understand deep {Q}-networks,'' \emph{IEEE Transactions on
  Visualization and Computer Graphics}, vol.~25, no.~1, pp. 288--298, 2019.

\bibitem{ming2017understanding}
Y.~Ming, S.~Cao, R.~Zhang, Z.~Li, Y.~Chen, Y.~Song, and H.~Qu, ``Understanding
  hidden memories of recurrent neural networks,'' in \emph{Proceedings of the
  IEEE Conference on Visual Analytics Science and Technology}, 2017, pp.
  13--24.

\bibitem{strobelt2018seq}
H.~Strobelt, S.~Gehrmann, M.~Behrisch, A.~Perer, H.~Pfister, and A.~M. Rush,
  ``Seq2seq-{Vis}: A visual debugging tool for sequence-to-sequence models,''
  \emph{IEEE Transactions on Visualization and Computer Graphics}, vol.~25,
  no.~1, pp. 353--363, 2019.

\bibitem{Liu2018Visual}
S.~Liu, J.~Xiao, J.~Liu, X.~Wang, J.~Wu, and J.~Zhu, ``Visual diagnosis of tree
  boosting methods,'' \emph{IEEE Transactions on Visualization and Computer
  Graphics}, vol.~24, no.~1, pp. 163--173, 2018.

\bibitem{Schneider2021}
B.~Schneider, D.~Jackle, F.~Stoffel, A.~Diehl, J.~Fuchs, and D.~Keim,
  ``Integrating data and model space in ensemble learning by visual
  analytics,'' \emph{{IEEE} Transactions on Big Data}, vol.~7, no.~3, pp.
  483--496, 2021.

\bibitem{Zhao2019}
X.~Zhao, Y.~Wu, D.~L. Lee, and W.~Cui, ``{iForest}: Interpreting random forests
  via visual analytics,'' \emph{{IEEE} Transactions on Visualization and
  Computer Graphics}, vol.~25, no.~1, pp. 407--416, 2019.

\bibitem{Neto2021}
M.~P. Neto and F.~V. Paulovich, ``Explainable matrix - {Visualization} for
  global and local interpretability of random forest classification
  ensembles,'' \emph{{IEEE} Transactions on Visualization and Computer
  Graphics}, vol.~27, no.~2, pp. 1427--1437, 2021.

\bibitem{chen2020oodanalyzer}
C.~Chen, J.~Yuan, Y.~Lu, Y.~Liu, H.~Su, S.~Yuan, and S.~Liu, ``{OoDAnalyzer}:
  Interactive analysis of out-of-distribution samples,'' \emph{IEEE
  Transactions on Visualization and Computer Graphics}, vol.~27, no.~7, pp.
  3335--3349, 2021.

\bibitem{yang2020diagnosing}
W.~Yang, Z.~Li, M.~Liu, Y.~Lu, K.~Cao, R.~Maciejewski, and S.~Liu, ``Diagnosing
  concept drift with visual analytics,'' in \emph{Proceedings of the IEEE
  Conference on Visual Analytics Science and Technology}, 2020, pp. 12--23.

\bibitem{ming2020protosteer}
Y.~Ming, P.~Xu, F.~Cheng, H.~Qu, and L.~Ren, ``{ProtoSteer}: Steering deep
  sequence model with prototypes,'' \emph{IEEE Transactions on Visualization
  and Computer Graphics}, vol.~26, no.~1, pp. 238--248, 2020.

\bibitem{gou2021vatld}
L.~Gou, L.~Zou, N.~Li, M.~Hofmann, A.~K. Shekar, A.~Wendt, and L.~Ren,
  ``{VATLD}: A visual analytics system to assess, understand and improve
  traffic light detection,'' \emph{{IEEE} Transactions on Visualization and
  Computer Graphics}, vol.~27, no.~2, pp. 261--271, 2021.

\bibitem{heimerl2012visual}
F.~Heimerl, S.~Koch, H.~Bosch, and T.~Ertl, ``Visual classifier training for
  text document retrieval,'' \emph{IEEE Transactions on Visualization and
  Computer Graphics}, vol.~18, no.~12, pp. 2839--2848, 2012.

\bibitem{behrisch2014feedback}
M.~Behrisch, F.~Korkmaz, L.~Shao, and T.~Schreck, ``Feedback-driven interactive
  exploration of large multidimensional data supported by visual classifier,''
  in \emph{Proceedings of the IEEE Conference on Visual Analytics Science and
  Technology}, 2014, pp. 43--52.

\bibitem{bruneau2013interactive}
P.~Bruneau and B.~Otjacques, ``An interactive, example-based, visual clustering
  system,'' in \emph{Proceedings of the International Conference on Information
  Visualisation}, 2013, pp. 168--173.

\bibitem{hoferlin2012inter}
B.~H{\"o}ferlin, R.~Netzel, M.~H{\"o}ferlin, D.~Weiskopf, and G.~Heidemann,
  ``Inter-active learning of ad-hoc classifiers for video visual analytics,''
  in \emph{Proceedings of the IEEE Conference on Visual Analytics Science and
  Technology}, 2012, pp. 23--32.

\bibitem{paiva2015approach}
J.~G.~S. Paiva, W.~R. Schwartz, H.~Pedrini, and R.~Minghim, ``An approach to
  supporting incremental visual data classification,'' \emph{IEEE Transactions
  on Visualization and Computer Graphics}, vol.~21, no.~1, pp. 4--17, 2015.

\bibitem{xiang2019interactive}
S.~Xiang, X.~Ye, J.~Xia, J.~Wu, Y.~Chen, and S.~Liu, ``Interactive correction
  of mislabeled training data,'' in \emph{Proceedings of the IEEE Conference on
  Visual Analytics Science and Technology}, 2019, pp. 57--68.

\bibitem{jia2021towards}
S.~Jia, Z.~Li, N.~Chen, and J.~Zhang, ``Towards visual explainable active
  learning for zero-shot classification,'' \emph{IEEE Transactions on
  Visualization and Computer Graphics}, vol.~28, no.~1, pp. 791--801, 2022.

\bibitem{elhamifar2015dissimilarity}
E.~Elhamifar, G.~Sapiro, and S.~S. Sastry, ``Dissimilarity-based sparse subset
  selection,'' \emph{IEEE Transactions on Pattern Analysis and Machine
  Intelligence}, vol.~38, no.~11, pp. 2182--2197, 2015.

\bibitem{schrijver1998theory}
A.~Schrijver, \emph{Theory of linear and integer programming}.\hskip 1em plus
  0.5em minus 0.4em\relax John Wiley \& Sons, 1998.

\bibitem{zhu2009introduction}
X.~Zhu and A.~B. Goldberg, \emph{Introduction to semi-supervised
  learning}.\hskip 1em plus 0.5em minus 0.4em\relax Morgan \& Claypool
  Publishers, 2009.

\bibitem{yang2015multi}
Y.~Yang, Z.~Ma, F.~Nie, X.~Chang, and A.~G. Hauptmann, ``Multi-class active
  learning by uncertainty sampling with diversity maximization,''
  \emph{International Journal of Computer Vision}, vol. 113, no.~2, pp.
  113--127, 2015.

\bibitem{Dinkla2012}
K.~Dinkla, M.~A. Westenberg, and J.~J. van Wijk, ``Compressed adjacency
  matrices: Untangling gene regulatory networks,'' \emph{{IEEE} Transactions on
  Visualization and Computer Graphics}, vol.~18, no.~12, pp. 2457--2466, 2012.

\bibitem{sneath1973numerical}
P.~H. Sneath, R.~R. Sokal \emph{et~al.}, \emph{Numerical taxonomy : the
  principles and practice of numerical classification}.\hskip 1em plus 0.5em
  minus 0.4em\relax W.H. Freeman, 1973.

\bibitem{pennington2014glove}
J.~Pennington, R.~Socher, and C.~D. Manning, ``{GloVe}: Global vectors for word
  representation,'' in \emph{Proceedings of the Conference on Empirical Methods
  in Natural Language Processing}, 2014, pp. 1532--1543.

\bibitem{liu2016visualizing}
S.~Liu, D.~Maljovec, B.~Wang, P.-T. Bremer, and V.~Pascucci, ``Visualizing
  high-dimensional data: Advances in the past decade,'' \emph{IEEE Transactions
  on Visualization and Computer Graphics}, vol.~23, no.~3, pp. 1249--1268,
  2017.

\bibitem{meng2015clutter}
Y.~Meng, H.~Zhang, M.~Liu, and S.~Liu, ``Clutter-aware label layout,'' in
  \emph{Proceedings of the IEEE Pacific Visualization Symposium}, 2015, pp.
  207--214.

\bibitem{mayorga2013splatterplots}
A.~Mayorga and M.~Gleicher, ``Splatterplots: Overcoming overdraw in scatter
  plots,'' \emph{{IEEE} Transactions on Visualization and Computer Graphics},
  vol.~19, no.~9, pp. 1526--1538, 2013.

\bibitem{yuan2020evaluation}
J.~Yuan, S.~Xiang, J.~Xia, L.~Yu, and S.~Liu, ``Evaluation of sampling methods
  for scatterplots,'' \emph{{IEEE} Transactions on Visualization and Computer
  Graphics}, vol.~27, no.~2, pp. 1720--1730, 2021.

\bibitem{endert2016semantic}
A.~Endert, ``Semantic interaction for visual analytics: inferring analytical
  reasoning for model steering,'' \emph{Synthesis Lectures on Visualization},
  vol.~4, no.~2, pp. 1--99, 2016.

\bibitem{zhou2012ensemble}
Z.-H. Zhou, \emph{Ensemble methods: {F}oundations and algorithms}.\hskip 1em
  plus 0.5em minus 0.4em\relax CRC press, 2012.

\bibitem{vinyals2016matching}
O.~Vinyals, C.~Blundell, T.~Lillicrap, K.~Kavukcuoglu, and D.~Wierstra,
  ``Matching networks for one shot learning,'' in \emph{Proceedings of the
  Advances in Neural Information Processing Systems}, 2016, pp. 3637--3645.

\bibitem{ren2018meta}
M.~Ren, E.~Triantafillou, S.~Ravi, J.~Snell, K.~Swersky, J.~B. Tenenbaum,
  H.~Larochelle, and R.~S. Zemel, ``Meta-learning for semi-supervised few-shot
  classification,'' in \emph{Proceedings of the International Conference on
  Learning Representations}, 2018.

\bibitem{lecun1998gradient}
Y.~LeCun, L.~Bottou, Y.~Bengio, and P.~Haffner, ``Gradient-based learning
  applied to document recognition,'' \emph{Proceedings of the IEEE}, vol.~86,
  no.~11, pp. 2278--2324, 1998.

\bibitem{bertinetto2018metalearning}
L.~Bertinetto, J.~F. Henriques, P.~Torr, and A.~Vedaldi, ``Meta-learning with
  differentiable closed-form solvers,'' in \emph{Proceedings of International
  Conference on Learning Representations}, 2019.

\bibitem{deng2009imagenet}
J.~Deng, W.~Dong, R.~Socher, L.-J. Li, K.~Li, and L.~Fei-Fei, ``{ImageNet}: A
  large-scale hierarchical image database,'' in \emph{Proceedings of the IEEE
  Conference on Computer Vision and Pattern Recognition}, 2009, pp. 248--255.

\bibitem{lake2015human}
B.~M. Lake, R.~Salakhutdinov, and J.~B. Tenenbaum, ``Human-level concept
  learning through probabilistic program induction,'' \emph{Science}, vol. 350,
  no. 6266, pp. 1332--1338, 2015.

\bibitem{Boudiaf2020TIM}
M.~Boudiaf, I.~Ziko, J.~Rony, J.~Dolz, P.~Piantanida, and I.~Ben~Ayed,
  ``Information maximization for few-shot learning,'' in \emph{Proceedings of
  the Advances in Neural Information Processing Systems}, 2020, pp. 2445--2457.

\bibitem{eck2009normalize}
N.~J.~v. Eck and L.~Waltman, ``How to normalize co-occurrence data? {An}
  analysis of some well-known similarity measures,'' \emph{Journal of the
  American Society for Information Science and Technology}, vol.~60, no.~8, pp.
  1635--1651, 2009.

\bibitem{arias2011pair}
R.~Arias-Hernandez, L.~T. Kaastra, T.~M. Green, and B.~Fisher, ``Pair
  analytics: Capturing reasoning processes in collaborative visual analytics,''
  in \emph{Proceedings of the International Conference on System Sciences},
  2011, pp. 1--10.

\bibitem{settles2009active}
B.~Settles, ``Active learning literature survey,'' University of
  Wisconsin-Madison, Tech. Rep., 2009.

\bibitem{fekete:hal-01361430}
J.-D. Fekete and R.~Primet, ``{Progressive Analytics: A Computation Paradigm
  for Exploratory Data Analysis},'' Jul. 2016.

\bibitem{fekete:hal-01202901}
J.-D. Fekete, ``{ProgressiVis: a Toolkit for Steerable Progressive Analytics
  and Visualization},'' in \emph{{1st Workshop on Data Systems for Interactive
  Analysis}}, Chicago, United States, Oct. 2015, p.~5.

\bibitem{williams2004steerable}
M.~Williams and T.~Munzner, ``Steerable, progressive multidimensional
  scaling,'' in \emph{Proceedings of the IEEE Symposium on Information
  Visualization}, 2004, pp. 57--64.

\bibitem{yi2007toward}
J.~S. Yi, Y.~ah~Kang, J.~Stasko, and J.~A. Jacko, ``Toward a deeper
  understanding of the role of interaction in information visualization,''
  \emph{{IEEE} Transactions on Visualization and Computer Graphics}, vol.~13,
  no.~6, pp. 1224--1231, 2007.

\bibitem{heer2012interactive}
J.~Heer and B.~Shneiderman, ``Interactive dynamics for visual analysis,''
  \emph{Communications of the ACM}, vol.~55, no.~4, pp. 45--54, 2012.

\end{thebibliography}
